\documentclass{SCIS2026}

\usepackage{amsmath,amssymb,amsfonts,bm}
\usepackage{graphicx}
\usepackage{booktabs}
\usepackage{tabularx}
\usepackage{array}
\newcolumntype{Y}{>{\raggedright\arraybackslash}X}
\usepackage{adjustbox}
\usepackage{caption}
\usepackage{multirow}
\usepackage{float}
\usepackage{hyperref}
\hypersetup{colorlinks=true, linkcolor=blue, citecolor=blue, urlcolor=blue}
\newcommand{\secref}[1]{\hyperref[#1]{Sec.~\ref*{#1}}}

\usepackage{amsthm}
\makeatletter
\@ifundefined{definition}{\newtheorem{definition}{Definition}}{}
\@ifundefined{proposition}{}{}
\makeatother
\newcommand{\aeqsbincludegraphics}[2][]{%
  \IfFileExists{#2}{\includegraphics[#1]{#2}}{%
    \fbox{\parbox[c][0.28\textheight][c]{0.86\linewidth}{\centering\small Figure placeholder: \texttt{\detokenize{#2}}}}%
  }%
}

\begin{document}

\ArticleType{RESEARCH PAPER}
\Year{2026}
\Month{June}
\Vol{69}
\No{}
\DOI{}
\ArtNo{}
\ReceiveDate{}
\ReviseDate{}
\AcceptDate{}
\OnlineDate{}
\AuthorMark{Liu D M, et al.}
\AuthorCitation{Liu D M, Li J, Chen X B, Wang J T. Adaptive Enhanced Quantum-inspired Simulated Bifurcation Algorithm for Population State Perception.}

\title{Adaptive Enhanced Quantum-inspired Simulated Bifurcation Algorithm for Population State Perception}

\author[1]{Dongmei LIU}{liudomain@163.com}
\author[1\dag]{Jian LI}{Lijian@bupt.edu.cn}
\author[1]{Xiubo CHEN}{}
\author[1]{Jin-Tao WANG}{}

\address[1]{State Key Laboratory of Networking and Switching Technology,\protect\\ Cyber Security Center, School of Cyberspace Security,\protect\\ Beijing University of Posts and Telecommunications, Beijing 100876, China}

\abstract{
Existing quantum-inspired simulated bifurcation algorithms rely on dynamic scheduling methods but lack the ability to adapt effectively to different problem instances. Additionally, during the evolutionary stage, balancing exploration and exploitation remains challenging. The fundamental issue stems from the widespread use of static preset parameters and globally uniform strategies, which can diminish algorithm effectiveness and lead to result homogenization. This article proposes an Adaptive Enhanced Quantum-inspired Simulated Bifurcation (AE-QSB) framework driven by population states. By leveraging perception indicators of four distinct population states, the QSB algorithm establishes a closed-loop strategy encompassing perception, decision-making, and execution. Within this framework, we introduce three complementary algorithms spanning a spectrum from efficient extremum seeking (ME-BSB), through population-level uniform refinement (SE-DSB), to density-aware adaptive scheduling (SG-DSB). On the medium-sized graph G22, both SE-DSB and SG-DSB achieve a mean gap below 0.05\%, while ME-BSB attains the optimal trade-off between runtime and solution quality with a gap of 0.26\% and the shortest single-run time. We compared AE-QSB variants with other algorithms across all benchmark graphs from G1 to G81. The results demonstrate that AE-QSB achieved the lowest mean gap on 74.6\% of the graphs and the highest average approximation rate on 84.5\% of the graphs. Ablation experiments further revealed that subgroup exploration and rescue mechanisms play crucial roles in both multifactor and single-factor components. This study demonstrates that population statistical information during dynamic evolution provides a computable and effective foundation for adaptive control, enabling quantum-inspired optimization methods to transition from fixed scheduling to data-driven closed-loop control.
}


\maketitle

\section{Introduction}
\label{sec:intro}
The combinatorial optimization problem is prevalent in various fields, including communication network design, computational biology, financial engineering, drug discovery, chip layout, and scientific computing. It represents one of the core challenges in information science and engineering optimization~\cite{jiQuantuminspiredGenericOptimization2026,jiangQuantuminspiredBeamformingOptimization2025,wangVariationalQuantumEigensolver2025,jangNextGenerationGraph2025,okawaQuantumannealinginspiredAlgorithmsMultijet2025,okawaQuantumannealinginspiredAlgorithmsTrack2024,furiniQPLIBLibraryQuadratic2019}. From the perspective of computational complexity, problems such as Maxcut, minimum vertex cover, graph coloring, Boolean satisfiability, portfolio optimization, and beamforming can all be formulated as Ising models.
 
\begin{equation}
\min_{\mathbf{s} \in \{\pm 1\}^N} \; H(\mathbf{s}) = -\frac{1}{2}\sum_{i,j} J_{ij} s_i s_j - \sum_i h_i s_i,
\label{eq:ising}
\end{equation}
where $s_i \in \{\pm 1\}$ denotes the spin variable, $J_{ij}$ are the entries of the Ising coupling matrix $J = -W$, and $h_i$ is the local external field (complete sign conventions in \secref{sec:prelim}). These problems commonly exhibit high dimensionality, strong coupling, and significant non-convexity. In the energy landscape, the number of local minima increases exponentially with problem size, presenting a combinatorial explosion challenge for exact algorithms. Meanwhile, heuristic methods are sensitive to complex energy landscapes and tend to prematurely converge to suboptimal basins~\cite{aramonPhysicsinspiredOptimizationQuadratic2019,liuEnhancedOpensourceScatter2025}. Therefore, developing new optimization methods that integrate efficiency, scalability, and global search capabilities remains a central focus of research in this field.

Research on the combinatorial optimization problem is gradually shifting from traditional single-heuristic search methods to parallel, hardware-accelerated, interpretable, physics-inspired dynamics. This approach has fostered a rich research ecosystem, including digital annealing~\cite{kameyamaBenchmarksDigitalAnnealer2024}, simulated bifurcation quantum annealing~\cite{pawlowskiSimulatedBifurcationQuantum2026}, mean-field annealing~\cite{kingNMFAEmulatingCoherent2018,veszeliMeanFieldApproximation2021,gunathilakaMeanFieldCoherent2023}, local quantum-inspired annealers~\cite{bowlesQuadraticUnconstrainedBinary2022}, and various Ising machine frameworks based on optical and electronic implementations~\cite{tiunovAnnealingSimulatingCoherent2019,rahLowPowerCoherent2023,sankarBenchmarkingQuantumAlgorithms2024}. Among these, the Coherent Ising Machine (CIM) has attracted significant attention due to its parallel evolution characteristics and potential for physical implementation~\cite{yamamoto2020coherent,hamerlyExperimentalInvestigationPerformance2019,bohmUnderstandingDynamicsCoherent2018}. CIM maps the original problem onto an Ising model and searches for low-energy eigenstates within complex energy landscapes through nonlinear dynamical systems. Benchmark models such as Maxcut and Ising provide a unified platform for evaluating solution quality, computational efficiency, and scalability across different algorithms~\cite{furiniQPLIBLibraryQuadratic2019,jalowieckiBruteforcingSpinglassProblems2021,mcgeoch2019guide}. The Ising solution framework based on oscillatory networks demonstrates strong scalability and computational competitiveness, evolving from early systems with hundreds of spins to larger-scale systems comprising 2,000 nodes and even 100,000 spins ~\cite{mcmahon2016fully,inagaki2016coherent,honjo2021100000,yamaoka201620k}.

The Simulated Bifurcation (SB) algorithm serves as a crucial link between the nonlinear dynamics of  CIM and general combinatorial optimization solutions~\cite{goto2016bifurcation,gotoSB_CombinatorialOptimizationSimulating2019}. Unlike approaches that depend on specialized quantum hardware, SB can be implemented on classical hardware platforms such as FPGAs and GPUs while preserving quantum-inspired search capabilities, thereby significantly lowering the barrier to adoption~\cite{tatsumura2019fpga,tatsumuraLargescaleCombinatorialOptimization2021,tatsumura2021scaling,zhangHighperformanceStochasticSimulated2024,zhang2025qsbm,volpe2025improving}. Compared to other heuristic methods, SB leverages the dynamic behavior near the bifurcation critical point to improve its ability to explore complex energy landscapes ~\cite{gotoEdgeofchaosenhancedQuantuminspiredAlgorithm2026}.

SB continues to evolve in four key directions: hardware scaling, dynamic enhancement, guided search, and benchmark algorithm evaluation. Regarding hardware, Goto et al. demonstrated that SB based on classical mechanics can achieve high-performance combinatorial optimization on FPGA~\cite{goto2016bifurcation,goto2021high}. Volpe et al. developed an open-source digital architecture~\cite{volpe2025improving}. Tatsumura et al. designed a large-scale FPGA acceleration scheme for real-time systems and extended it to multi-chip architectures with millions of variables, also applying it to FPGA-based currency trading systems~\cite{tatsumuraLargescaleCombinatorialOptimization2021,tatsumura2021scaling,tatsumura2019fpga,tatsumura2025currency}. Zhang et al. proposed quantitative SB Ising machines (QSBMs)~\cite{zhangHighperformanceStochasticSimulated2024,zhang2025qsbm}. In terms of dynamic enhancement, Leleu et al. introduced Chaos Amplitude Control (CAC), which stabilizes local minima by correcting amplitude heterogeneity, subsequently demonstrating CAC’s scaling advantage on large benchmarks~\cite{leleuDestabilizationLocalMinima2019,leleuScalingAdvantageChaotic2021}. Wang et al. analyzed the bifurcation behavior mechanism in solving discrete Ising optimization problems within continuous physical dynamics from the perspective of bifurcation theory~\cite{wangBifurcationBehaviorsShape2023}. Kanao and Goto explicitly incorporated thermal fluctuations into SB dynamics~\cite{kanaoSimulatedBifurcationAssisted2022}. Goto et al. explored the parameter region of chaotic edges~\cite{gotoEdgeofchaosenhancedQuantuminspiredAlgorithm2026}. Lee et al. systematically studied the optimization and sampling capabilities of noise-enhanced chaotic Ising machines~\cite{lee2025noise}. Regarding guided search, Xiao et al. proposed Global Guided SB (GSB), introducing a comprehensive learning mechanism involving gbest tracking and velocity momentum memory~\cite{xiaoGloballyGuidedSimulated2026}. Tao et al.’s Tabu-SB steers the search away from known suboptimal regions through repulsive fields~\cite{taoTabuenhancedSimulatedBifurcation2026}. Shen et al.’s free energy machine unified multiple Ising solving methods from the perspective of minimizing free energy in statistical physics~\cite{shen2024free}. Finally, in benchmark evaluation, Zeng et al. systematically compared quantum-inspired algorithms—including SB variants, CIM simulations, and simulated annealing—on G-set and QPLIB datasets~\cite{zeng2024performance}.

Although significant progress has been made in the aforementioned SB enhancement methods, most still rely on fixed parameter scheduling. The noise amplitude, chaotic disturbance intensity, or amplitude correction typically remains constant throughout the entire evolution process or gradually decays according to a preset function. This approach makes it difficult to perform real-time adjustments based on different problem instances and search states at various stages of evolution. Consequently, existing methods have not fundamentally established a closed-loop adaptive control mechanism that focuses on operational state feedback and graph structure differences. In contrast, adaptive parameter control emphasizes dynamically adjusting algorithm behavior based on feedback information during the search process, which is highly significant for complex optimization problems~\cite{fuEvolutionaryIsingOptimization2025,ouyangLearnGlobalCorrelations}. However, how to utilize population state information to achieve online control within the SB dynamics framework remains an open problem requiring further research. Building on this concept, AE-QSB incorporates runtime population statistics into the SB evolution process. The entire process involves real-time monitoring of population diversity, freezing degree, symbol-flipping activity, and recent improvements, dynamically adjusting coupling modes, step sizes, elite guidance, and restart strategies. This design principle can also serve as a reference for other parallel Ising solving frameworks.
 
The limitations of existing SB methods are particularly evident in three typical scenarios. First, uniform guidance can easily lead to population homogenization. For example, GSB employs a constant gbest guidance intensity~\cite{xiaoGloballyGuidedSimulated2026}, which can cause a rapid loss of population diversity in G22 instances and premature clustering of candidate solutions within the same suboptimal basin. Second, fixed stage division can result in scheduling mismatches. The hard-coded two-stage strategy of Tabu-SB~\cite{taoTabuenhancedSimulatedBifurcation2026} struggles to adapt to varying convergence rates across different instances, potentially wasting iteration budgets on fast-converging graphs or prematurely entering the refinement stage on slow-converging graphs. Third, noise or chaos enhancement mechanisms still depend on manual parameter tuning. Methods involving thermal fluctuations and chaotic edges~\cite{kanaoSimulatedBifurcationAssisted2022,gotoEdgeofchaosenhancedQuantuminspiredAlgorithm2026} typically require case-by-case adjustment of disturbance intensity, timing, and decay rate, lacking an automatic feedback mechanism driven by the system’s operating state. Collectively, these issues suggest that improving SB performance necessitates a shift from open-loop static scheduling to closed-loop, population-state-driven adaptive control.

In response to the aforementioned shortcomings, this paper proposes the AE-QSB (Adaptive Enhanced Quantum-inspired Simulated Bifurcation) framework, which integrates a real-time state perception mechanism of the population into the SB dynamic evolution process and drives adaptive control through runtime feedback. The core concept is that population statistics can characterize key evolutionary states during the search process and provide computable, interpretable feedback signals for dynamically adjusting algorithm behavior. AE-QSB introduces a lightweight perception layer within batch SB evolution, calculating four complementary metrics $(D, F, Q, R)$ every $\Delta_{\text{eval}}$ step to characterize symbol space diversity, amplitude discretization, symbol flipping activity, and recent improvements in the optimal value. Based on these state variables, AE-QSB dynamically adjusts the coupling mode, candidate solution step size, elite guidance strength, restart strategy, and early stopping conditions, forming a closed "perception-decision-execution" loop. This framework is decoupled from specific SB implementations and can serve as a universal enhancement layer adaptable to different SB-based solvers. 

Under this framework, we have designed algorithm examples for two complementary routes. ME-BSB (Mode-switching Enhanced Ballistic to Discrete SB, \secref{sec:me-bsb}) adopts an $F$- driven BSB-to-DSB hard switching strategy. This approach combines elite guidance with a 15\% weak search subgroup to maintain independent search freedom, embodying the design principle of "exploration before commitment." The gradient information from the early BSB continuous coupling enables highly efficient basin discovery; however, the depth of refinement is limited after the hard switch to DSB. SE-DSB and SG-DSB (\secref{sec:se-sg-dsb}) form variant pairs of the DSB route: SE-DSB (Static Enhanced Discrete SB) serves as the baseline architecture, employing linear mixed coupling combined with a gate control rescue mechanism. SG-DSB (Smoothed Guided Discrete SB) builds on this by introducing density-aware adaptive parameter scheduling and velocity momentum smoothing, enabling the algorithm to adapt its behavior across different graph densities. Both methods achieved a performance gap of approximately 0.04\%–0.05\% on G22, and the system ablation experiment (\ref{app:ablation}) confirmed the marginal contributions and interaction effects of each component.

This article provides a concise summary across four levels. At the framework level, it strictly defines and analyzes four population state perception indicators $(D, F, Q, R)$, demonstrating their composition as sufficient and non-redundant evolutionary state representations, and constructs a closed-loop adaptive architecture (\secref{sec:framework}).  At the algorithm level, three algorithm examples—ME-BSB, SE-DSB, and SG-DSB—are proposed (\secref{sec:me-bsb}-\secref{sec:se-sg-dsb}), introducing simplified mechanisms such as single elite guidance, gated assistance, linear mixed coupling with smooth transition, and density-aware adaptive scheduling.  At the experimental level, statistical validation was conducted 10 times on two main benchmark sets, G22 and G1–G81 (\secref{sec:setup}-\secref{sec:ttsar}): the mean gap of SE-DSB and SG-DSB was lower than that of other algorithms, while ME-BSB achieved the optimal balance between time and quality. The AE-QSB variant attained the lowest mean gap and highest mean approximation rate on most graphs.  At the ablation methods level, a dual-layer ablation system combining single-factor and multi-factor combinations was established to reveal the super-additive complementarity and functional redundancy among components (\ref{app:ablation}).

Paper Organization: \secref{sec:prelim} introduces SB Dynamics; \secref{sec:framework} elaborates on the AE-QSB framework; \secref{sec:experiments} presents algorithm examples and reports experimental results for G22 and G1–G81; \secref{sec:discussion} discusses ablation findings, algorithm complementarity, and limitations; \secref{sec:conclusion} provides a summary of the entire paper.

\section{Preliminaries}
\label{sec:prelim}

\subsection{Problem formulation and notation}
\label{sec:ising}

Given an undirected weighted graph $\mathcal{G} = (\mathcal{V}, \mathcal{E}, W)$ with $|\mathcal{V}| = N$ and edge weights $w_{ij} \in \mathbb{R}$, a cut is defined by a spin configuration $\mathbf{s} \in \{\pm 1\}^N$. Its value is:
\begin{equation}
C(\mathbf{s}) = \frac{1}{2}\sum_{i<j} w_{ij}(1 - s_i s_j).
\label{eq:cut}
\end{equation}
The Maxcut problem seeks $\max_{\mathbf{s}} C(\mathbf{s})$. Defining $J = -W$ and $W_{\text{total}} = \sum_{i<j} w_{ij}$, one obtains:
\begin{equation}
C(\mathbf{s}) = \frac{2W_{\text{total}} + \mathbf{s}^\top J \mathbf{s}}{4}.
\label{eq:cut-mat}
\end{equation}
$W_{\text{total}}$ is a fixed global constant, and the energy term is $\mathbf{s}^\top J \mathbf{s}$, where the batch cut value $C(s)$ can be computed with a single sparse matrix multiplication. If there exists a spin assignment $\mathbf{s} = \{s_1, \dots, s_N\} \in \{\pm 1\}^N$ such that all edge constraints simultaneously satisfy $s_i s_j = \operatorname{sgn}(J_{ij}), \forall (i,j) \in \mathcal{E}$, the graph $\mathcal{G}$ is \textit{unfrustrated}. The following unified notation is adopted; key symbols are summarized in Table~\ref{tab:notation}.

\begin{table}[!t]
\centering
\caption{Notation conventions.}
\label{tab:notation}
\scriptsize
\begin{adjustbox}{max width=\textwidth}
\begin{tabular}{lll}
\toprule
Symbol & Definition & Typical value \\
\midrule
$N$ & Number of graph vertices & 800-20000 \\
$J$ & Ising coupling matrix ($J = -W$) & - \\
$B$ & Batch size (candidate columns) & 256 \\
$T$ & Total iteration count & 1000-5000 \\
$\tau = t/T$ & Normalized time & $[0, 1]$ \\
$a(t)$ & Time-varying bifurcation parameter & $[0, 1]$ \\
$\xi$ & Coupling strength scaling factor & Eq.~\eqref{eq:xi} \\
$\mu$ & Base step size & 0.5-1.8 \\
$\Delta_{\text{eval}}$ & Sensing evaluation interval & 50 steps \\
$\mathbf{x}, \mathbf{y}$ & Amplitude and momentum matrices ($\mathbb{R}^{N \times B}$) & - \\
$\mathbf{s}$ & Discrete spins ($\mathbf{s} = \operatorname{sgn}(\mathbf{x})$) & $\{\pm 1\}^{N \times B}$ \\
\bottomrule
\end{tabular}
\end{adjustbox}
\end{table}

\subsection{SB dynamics and batch parallelism}
\label{sec:sb-dyn}

The SB dynamics originate from the supercritical pitchfork bifurcation of Kerr-nonlinear oscillator networks~\cite{goto2016bifurcation,gotoSB_CombinatorialOptimizationSimulating2019}. In the early evolution, continuous amplitude variables provide gradient-like global exploration capability, helping the system search for high-quality basins in complex energy landscapes; as amplitudes gradually saturate, variable signs stabilize, and the algorithm transitions into terminal refinement in discrete spin space. The continuous-time evolution is governed by Hamiltonian dynamics:
\begin{equation}
\begin{aligned}
\dot{x}_i &= a(0) \cdot y_i, \\
\dot{y}_i &= -\bigl(a(0) - a(t)\bigr) x_i + \xi \sum_{j} J_{ij}\, \phi(x_j),
\end{aligned}
\label{eq:sb-cont}
\end{equation}
Standard SB typically adopts a linear schedule $a(t) = \tau$ with $\tau = t/T$. AE-QSB employs a nonlinear schedule $a(t) = a(0)\cdot\tau^\gamma$ (see \secref{sec:decision}). Euler discretization of Eq.~\eqref{eq:sb-cont} with step size $\mu$ yields the batch-parallel iterative form:
\begin{equation}
\begin{aligned}
\mathbf{y}^{(t+1)} &= \mathbf{y}^{(t)} + \Bigl[-\bigl(a(0) - a(t)\bigr) \mathbf{x}^{(t)} + \xi \cdot J \boldsymbol{\phi}(\mathbf{x}^{(t)})\Bigr] \mu, \\
\mathbf{x}^{(t+1)} &= \mathbf{x}^{(t)} + a(0) \cdot \mu \cdot \mathbf{y}^{(t+1)}, \\
(x_i, y_i) &\leftarrow (\operatorname{sgn}(x_i), 0) \quad \text{if } |x_i| > 1.
\end{aligned}
\label{eq:sb-disc}
\end{equation}

Depending on the coupling method, SB can generally be categorized into three modes: BSB (Ballistic SB), DSB (Discrete SB), and hybrid coupling. BSB employs continuous amplitude coupling, defined as $\phi(x) = x$, which preserves amplitude information during the early stages. This ensures a smoother evolution process and facilitates crossing narrow energy barriers. However, as BSB approaches the final state, continuous amplitude fluctuations may introduce additional gradient noise, thereby reducing convergence efficiency. DSB utilizes discrete sign coupling, $\phi(x) = \operatorname{sgn}(x)$,  which aligns more closely with discrete Ising targets. DSB provides a more direct discrete optimization direction; however, if applied too early, it may lose valuable amplitude information, weakening global exploration capabilities. To stabilize dynamic evolution across graphs of varying scales, the coupling strength $\xi$ adopts a variance-normalized form:

\begin{equation}
\xi = \frac{0.5\sqrt{N-1}}{\sqrt{\sum_{i,j} J_{ij}^2}}.
\label{eq:xi}
\end{equation}

In the batch parallel implementation, $B $candidate solutions form a population represented as columns in a matrix and evolve simultaneously. Denote the amplitude vector of column $b$ ($b = 1, \dots, B$) as $\mathbf{x}_b \in \mathbb{R}^N$. The corresponding cut value is $C_b = C(\operatorname{sgn}(\mathbf{x}_b))$. The overall quality of the population should be evaluated not only by the performance of the best individual candidate solutions but also by the statistical distribution of all $C_b $ values. A high-quality population should exhibit both a high average of $C_b $ and sufficient diversity to prevent premature convergence to a single configuration during evolution.
\section{AE-QSB: Population state-aware adaptive enhancement framework}
\label{sec:framework}

\subsection{Theoretical basis}
\label{sec:design}
The core of AE-QSB is to perform statistical analysis of population state information in batches through algorithmic control logic during the SB evolution process. Unlike traditional fixed scheduling, AE-QSB employs real-time population state information to drive closed-loop adaptive control. In the parallel evolution of $B$ candidate solutions, the SB solver must not only explore the initial state multiple times randomly but also consider factors such as the search stage, convergence status, and stagnation risk. Therefore, AE-QSB transcends the limitation of focusing solely on a single optimal individual target value by leveraging the structural characteristics of the entire population as a key criterion to determine whether the current algorithm is in the exploration, transition, freezing, or stagnation stage.

In AE-QSB, the population is characterized by four indicators—$D, F, Q$, and $R$—each representing distinct state information. Diversity $D$ measures the degree of dispersion within the population in symbol space and is used to detect premature convergence. The freezing rate $F$ quantifies the proportion of continuous amplitude variables approaching saturation boundaries, indicating whether the population has entered the discretization phase. The flipping rate $Q$ reflects the frequency of symbol configuration changes between adjacent evaluation windows, capturing dynamic activity; it also informs adaptive bit-flip operations (Eq.~\ref{eq:bitflip}) and early stopping criteria. The improvement rate $R$ denotes the relative increase in the objective function within the most recent window, tracking optimization progress. These four indicators correspond to population information across four dimensions: symbol structure, amplitude saturation, dynamic activity, and target improvement. Each indicator complements the others without redundancy. Experimental results demonstrate that relying on any single indicator makes it difficult to reliably distinguish evolutionary states such as normal convergence, premature collapse, active search, and ineffective oscillation. Therefore, AE-QSB employs a multi-index joint discrimination mechanism to achieve more effective adaptive control.

Based on the state representation described above, we have developed an AE-QSB closed-loop framework comprising three levels: perception, decision, and execution. The perception layer calculates the four population state indicators mentioned earlier at every $\Delta_{\text{eval}}$ step. The decision-making layer dynamically adjusts the step size and coupling mode based on these indicators to implement elite guidance, restart, and early stopping strategies. The execution layer translates the decision outcomes into parameter updates or population operations within SB dynamics. This framework has two key characteristics: it is lightweight and decoupled. The lightweight nature is reflected in the computational complexity of the four indicators, which is $O(NB)$, and the fact that these indicators do not involve matrix operations during the $\Delta_{\text{eval}}$ step, resulting in minimal impact on overall runtime. Decoupling means that the framework does not depend on specific SB variants and can be integrated into CAC, CIM, and other batch Ising solvers. Therefore, AE-QSB is designed as a state feedback enhancement mechanism for batch SB solvers within the scope of this work, with its metric definitions independent of particular SB variants.

\subsection{Population-state sensing indicators}
\label{sec:sensing}
The optimal cut value of the current window population, denoted as $C_{\text{best}} = \max_b C_b$ refers to the maximum cut value among the $B$ candidate solutions evaluated in the current step. $C_{\text{best}}$ is used to measure the best performance of the population within the most recent window and is applied in the improvement rate $R$ (Eq.~\ref{eq:R-def}) and the step size formula (Eq.~\ref{eq:mu-b}). The elite solution is a consistently high-quality configuration selected from the evolutionary history through Hamming distance filtering (Eq.~\ref{eq:ham-filter}). Its cut value may be slightly lower than the current $C_{\text{best}}$ because updating the elite pool requires the new solution to be sufficiently different from existing elite solutions, with a Hamming distance threshold of($d_H \geq 0.02$). In other words, the elite solution represents the "best after historical deduplication" rather than the "best in the current window." The worst solution in the current window, $C_{\text{worst}} = \min_b C_b$ is used only as the normalized denominator in the stride formula. The following sections provide a detailed introduction to these relevant indicators.
  
\begin{definition} Definition 1 (Diversity $D$). Diversity  $D$ characterizes the average degree of inconsistency within a population in the symbolic space, representing the uncertainty of the marginal distribution of variables. Let $s_{i,b}=\operatorname{sgn}(x_{i,b})$ represent the sign of the $b$-th candidate solution for the $i$-th variable, defined as:
\label{def:D}
\begin{equation}
D = 1 - \frac{1}{N}\sum_{i=1}^{N}\left|\frac{1}{B}\sum_{b=1}^{B} s_{i,b}\right| \in [0, 1].
\label{eq:D-def}
\end{equation}
For a single variable $i$, if $B$ candidate solutions tend to have consistent signs in that dimension, then $\left|\frac{1}{B}\sum_{b=1}^{B}s_{i,b}\right|$ is close to 1; consequently, the contribution of this term to $D$ is close to zero. Conversely, if the positive and negative signs are approximately balanced, then $\left|\frac{1}{B}\sum_{b=1}^{B}s_{i,b}\right|$ is close to 0, and its contribution to $D$ is close to 1. Essentially, $D$ represents the average anisotropy of the population in the $N$-dimensional symbol space: when $D$ approaches 1, the population maintains high symbol diversity and exhibits strong exploratory ability; when $D$ approaches 0, the population tends toward a single symbol configuration, which poses a risk of premature convergence.
\end{definition}

\begin{definition}Definition 2 (Freeze Rate $F$). The freeze rate $F$ measures the proportion of continuous amplitude variables approaching the saturation boundary, indicating whether the population has entered the discretization stage. It is defined as follows:
\label{def:F}
\begin{equation}
F = \frac{1}{NB}\sum_{i=1}^{N}\sum_{b=1}^{B} \mathbf{1}\bigl[|x_{i,b}| > 0.98\bigr] \in [0, 1].
\label{eq:F-def}
\end{equation}
Among them, $\mathbf{1}[\cdot]$ is the indicator function. When $|x_{i,b}| =1$, it indicates that the corresponding variable has reached a discrete sign state. An increase in $F$ signifies  that the amplitude of most variables in the population has reached saturation. At this point, the algorithm can continue to rely on continuous amplitude for global exploration. If the marginal returns decrease, the algorithm should promptly shift to discrete coupling or local bit-flip refinement, or trigger an early stopping criterion.

\end{definition}

\begin{definition}Definition 3 (Flip rate $Q$).  The flip $Q$ characterizes the frequency of symbol configuration changes between adjacent evaluation windows, measures the dynamic activity of the population, and is defined as:
\label{def:Q}
\begin{equation}
Q = \frac{1}{NB}\sum_{i=1}^{N}\sum_{b=1}^{B} \mathbf{1}\bigl[s_{i,b}^{(t)} \neq s_{i,b}^{(t - \Delta_{\text{eval}})}\bigr] \in [0, 1].
\label{eq:Q-def}
\end{equation}
Among these, $\Delta_{\mathrm{eval}}$ is the perceptual evaluation interval. A high value of $Q$ indicates that the candidate solution is still frequently flipping within the symbol space, implying that the dynamics remain active and the search has not yet converged. Conversely, a low $Q$ suggests that the symbol configuration is tending to stabilize. The parameter $Q$ serves two important functions in AE-QSB. First, the combination of $Q$ and $F$ forms the maturity level, defined as $F \cdot (1-Q)$, which adaptively determines the number of bits to flip (see Eq.~\ref{eq:bitflip}). A high freezing rate indicates amplitude saturation, while a low flipping rate confirms symbol stability; both are essential. Second,$Q$ is one of the criteria for early stopping. Termination is triggered only when $Q$ drops to an extremely low level, in conjunction with $R=0$ and $F>0.98$, effectively preventing collapse or pseudo-convergence of the GSB-BSB equation. In this scenario, $F$ approaches 1, but the sign is locked rather than naturally converging. $Q$ does not participate in step size adjustment (Eq.~\ref{eq:mu-b}) because it overlaps with $F$ in describing convergence.
\end{definition}
 
\begin{definition}Definition 4 (Improvement Rate $R$). The improvement rate $R$ quantifies the relative enhancement of the objective function over the most recent perception window and is defined as:
\label{def:R}
\begin{equation}
R = \operatorname{clip}\!\left(\frac{C_{\text{best}}^{(t)} - C_{\text{best}}^{(t - \Delta_{\text{eval}})}}{|C_{\text{best}}^{(t - \Delta_{\text{eval}})}| + \varepsilon},\; 0,\; 1\right),
\label{eq:R-def}
\end{equation}
Among these, $\varepsilon = 10^{-8}$ prevents division by zero. A positive value of $R > 0$ indicates that the target value within the recent window has improved, and the degree of improvement is quantified by the coefficient $(1 + \rho_R R)$, which is used to calculate the step size amplification factor (Eq.~\ref{eq:mu-b}). Conversely, if multiple consecutive windows have $R=0$, triggering the stall counter will result in an early stop.
\end{definition}

\subsection{Adaptive decision-making mechanisms}
\label{sec:decision}

The four sensing indicators drive five adaptive mechanisms. $D$ feeds into gated guidance (Eq.~\ref{eq:guidance}) and emergency restart; $F$ participates in step size (Eq.~\ref{eq:mu-b}), mode switching (Eq.~\ref{eq:mode}), and early stopping; $Q$ participates in bitflip (Eq.~\ref{eq:bitflip}) and early stopping; $R$ participates in step size (Eq.~\ref{eq:mu-b}) and early stopping. The instantiation design of each mechanism is described in \secref{sec:me-bsb}-\secref{sec:se-sg-dsb}.

\medskip\noindent\textbf{1. Column-by-column adaptive step size:} reduce the step size for columns approaching the elite solution to enable local refinement, and increase the step size for columns lagging behind to escape low-quality regions. The step size is determined by the improvement status, freezing degree, and relative solution quality.
\begin{equation}
\mu_b = \operatorname{clip}\!\left(\mu_0 \cdot (1 + \rho_R R) \cdot (1 - \rho_F F) \cdot \left(1 + \alpha_{\text{gap}}\frac{C_{\text{best}} - C_b}{C_{\text{best}} - C_{\text{worst}} + \varepsilon}\right),\;\mu_{\min},\;\mu_{\max}\right),
\label{eq:mu-b}
\end{equation}
The three factors each respond to different states: $(1+\rho_R R)$ amplifies the step size to accelerate convergence when significant improvement is achieved; $(1-\rho_F F)$ attenuates the step size to switch to a refinement phase when freezing intensifies; $(1+\alpha_{\text{gap}}\frac{C_{\text{best}}-C_b}{C_{\text{best}}-C_{\text{worst}}+\varepsilon})$ normalizes and scales the step size based on the difference from the optimal population. The elite solution maintains the base step size, while the worst solution increases it by approximately $1+\alpha_{\text{gap}}$ times to escape low-quality basins. The operator $\operatorname{clip}$ constrains the step size within $[\mu_{\min}, \mu_{\max}]$: $\mu_{\min}$ ensures minimum mobility, while $\mu_{\max}$ prevents excessively large step sizes that could cause frequent boundary violations. The parameter $Q$  does not participate in step size adjustment, as it overlaps with $F$ in describing convergence behavior.

\medskip\noindent\textbf{2. Coupling mode selection.} The transition from BSB ($\phi = \mathbf{x}$) to DSB ($\phi = \operatorname{sgn}(\mathbf{x})$) is jointly determined by $F$, $\tau$, and $\tau_{\min}$:
\begin{equation}
\phi = \begin{cases}
\mathbf{x}, & \tau < \tau_{\min}\; \lor \; F \leq \min\!\bigl(0.95,\;F_{\text{switch}} + \beta_{\text{dense}} \cdot (1 - \tau)\bigr), \\[4pt]
\operatorname{sgn}(\mathbf{x}), & \text{otherwise}.
\end{cases}
\label{eq:mode}
\end{equation}
The switching condition includes two lines of defense: $\tau \geq \tau_{\min}$ ensures that the evolution enters a stable phase before allowing switching. The dynamic threshold $\min(0.95,\,F_{\text{switch}} + \beta_{\text{dense}}(1-\tau))$ increases in the early stages due to the large value of $(1-\tau)$, which suppresses premature discretization, and gradually decreases to $F_{\text{switch}}$ as $\tau$ approaches 1 in the later stages. The upper limit of 0.95 prevents the threshold from becoming too strict.

\medskip\noindent\textbf{3. Diversity-Gated Elite Guidance.} The guidance vector directs the candidate solution toward the elite direction, with its strength determined by a combination of four factors:
\begin{equation}
\mathbf{g}_b = \alpha_{\text{gbest}} \cdot \omega(\tau) \cdot \min\!\left(\frac{D}{D_{\text{thresh}}}, 1\right) \cdot \xi \cdot (\mathbf{s}_k^{\text{elite}} - \mathbf{x}_b),
\label{eq:guidance}
\end{equation}
Among these, $\alpha_{\text{gbest}}$ represents the basic guidance strength (with ME-BSB fixed, and SE-DSB/SG-DSB density-aware dynamic adjustment). $\omega(\tau)$ denotes the stage scheduling: early preloading accelerates basin positioning, while mid-term regression benchmarking and sprint stage amplification enhance and refine the process. The term $\min(D/D_{\text{thresh}}, 1)$ acts as a self-stabilizing gating core that efficiently guides the process when $D$ exceeds the threshold, reduces the intensity by half when $D$ falls to half the threshold, and nearly closes when $D$ approaches zero. The guiding magnitude is normalized to the same scale as the coupling term. The vector $\mathbf{s}_k^{\text{elite}} - \mathbf{x}_b$ uses the elite symbol mode instead of continuous amplitude for directional guidance, thereby avoiding amplitude interference with directional information.

The gating factors described above form a negative feedback self-stabilization loop: when D is high, it exerts full influence, causing candidate solutions to converge toward the elite, which in turn reduces $D$ accordingly. Once $D$ falls below the threshold, the gate's decay guidance weakens, allowing $D$ to recover. In contrast, GSB~\cite{xiaoGloballyGuidedSimulated2026} lacks this mechanism and employs constant-intensity guidance, resulting in a positive feedback loop that leads to population collapse. Ablation confirms that a single elite solution outperforms multiple elites. We use an elite pool to maintain a single optimal solution ($K=1$), and new candidate solutions must satisfy the Hamming distance threshold to enter the elite.
\begin{equation}
d_H(\mathbf{s}_{\text{new}}, \mathbf{s}_k) = \frac{1}{N}\sum_{i=1}^{N} \mathbf{1}[s_{i,\text{new}} \neq s_{i,k}] \geq \theta_H = 0.02, \quad \forall k \in \{1, \dots, K\}.
\label{eq:ham-filter}
\end{equation}

\medskip\noindent\textbf{4. Emergency Restart and Elite Restart.} The restart mechanism serves as the last line of defense for maintaining population diversity. We employ two complementary approaches: emergency triggering and periodic injection. An emergency restart is triggered when  $D < 0.25$ and $\tau < 0.7$, directly re-initializing 30\% of the worst-performing columns in the population. The elite restart is triggered periodically at intervals of $T_{\text{restart}}$, replacing the worst column with an elite individual plus noise, and pushing it toward a tabu direction.
\begin{equation}
\mathbf{x}_{\text{bad}} = \mathbf{x}_{\text{elite}} + \sigma(\tau) \cdot \boldsymbol{\eta} - \beta \cdot \mathbf{h}_{\text{tabu}},
\label{eq:restart}
\end{equation}
The three terms above each have distinct characteristics: $\mathbf{x}_{\text{elite}}$ provides a high-quality starting point; $\sigma(\tau) \cdot \boldsymbol{\eta}$ (where $\boldsymbol{\eta} \sim \mathcal{N}(0, I)$) generates variants within the elite domain, with the noise intensity linearly attenuated according to the following formula:
\begin{equation}
\sigma(\tau) = \sigma_0 \cdot (1 - \tau) + \sigma_{\min},
\label{eq:sigma}
\end{equation}
When $\tau = 0$ in the early stage, the noise level is high, which facilitates the diffusion of variants outside the elite basin. At the late stage, when $\tau = 1$, the noise attenuates to its minimum value $\sigma_{\min}$ to focus on refinement. The taboo exclusion term $-\beta \cdot \mathbf{h}_{\text{tabu}}$ actively pushes the restart column away from known high-quality basins, thereby forcing the exploration of new areas in the energy landscape.

\medskip\noindent\textbf{5. Terminal refinement and early stopping.} Late in the evolution, adaptive bit-flip is applied to top solutions (Eq.~\ref{eq:bitflip}), with the number of flipped bits determined by maturity $F \cdot (1-Q)$. Early stopping triggers when $\text{stall} > 50 \;\land\; F > 0.98 \;\land\; Q < 0.05$; the triple condition ensures that improvement stagnation, amplitude saturation, and sign locking are simultaneously satisfied before termination.

\subsection{Unified framework}
\label{sec:execution}

\begin{figure}[!t]
\centering
\aeqsbincludegraphics[width=0.95\textwidth]{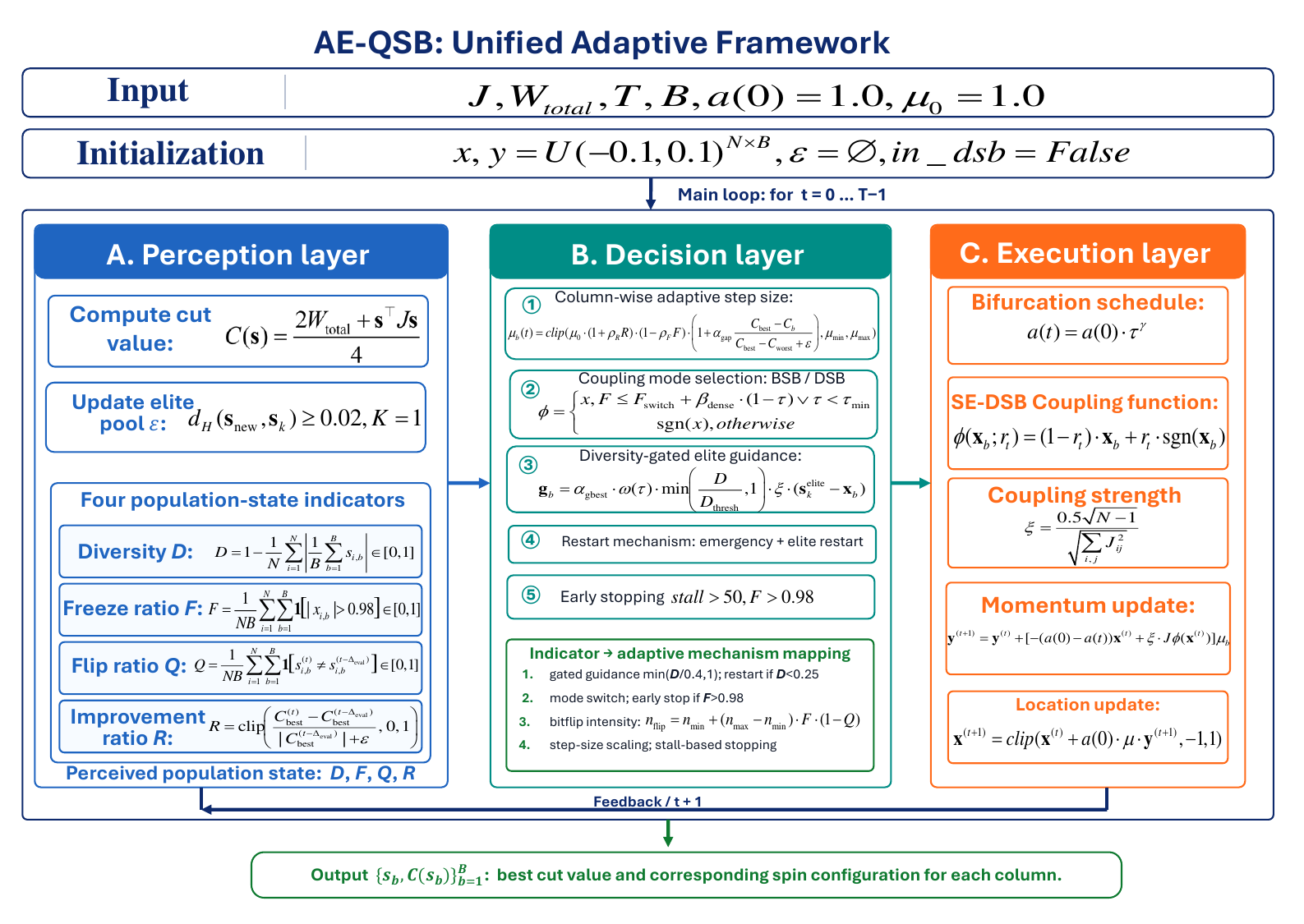}
\caption{AE-QSB unified adaptive framework. The perception layer calculates $(D, F, Q, R)$ at every $\Delta_{\text{eval}}$ step. The decision layer adjusts the step size, coupling mode, and guidance strength based on changes in these indicators. Finally, the execution layer performs dynamic updates, converting the decision layer's adjustments into the actual evolution of amplitude and momentum.}
\label{fig:framework}
\end{figure}

As shown in Figure \ref{fig:framework}, our algorithm's unified framework is presented. The five adaptive mechanisms of the decision layer affect the execution layer through the following paths. The step-size mechanism (Eq.~\ref{eq:mu-b}) outputs an independent step size $\mu_b$ for each column, constituting the per-column scaling factor in the momentum and amplitude updates. Coupling mode selection (Eq.~\ref{eq:mode}) determines the coupling function $\boldsymbol{\phi}$: BSB mode uses continuous amplitude $\mathbf{x}$, DSB mode uses discrete sign $\operatorname{sgn}(\mathbf{x})$, and SE-DSB/SG-DSB use a linear mixture of the two (Eq.~\ref{eq:mixed-coupling}). Diversity-gated elite guidance (Eq.~\ref{eq:guidance}) computes a guidance vector $\mathbf{g}_b$ for each column as an additional driving term in the momentum update. Emergency and elite restarts (Eqs.~\ref{eq:restart}-\ref{eq:sigma}) directly modify the amplitudes $\mathbf{x}$ and momenta $\mathbf{y}$ of the worst-performing columns. Early stopping terminates the main loop when conditions are met. The outputs of the five mechanisms are integrated at each step of the execution layer. Compared to the basic SB iteration (Eq.~\ref{eq:sb-disc}), the execution layer of AE-QSB introduces three enhancements: $(i)$ the vector $\boldsymbol{\phi}$ in the coupling term $J\boldsymbol{\phi}$ is dynamically selected by the decision layer (continuous \(\mathbf{x}\), discrete \(\operatorname{sgn}(\mathbf{x})\), or a linear mixture); $(ii)$ the elite guidance term $\mathbf{g}$ has been added to the momentum updates, with its strength regulated by diversity gating factors; $(iii)$ the global step size is replaced by a column-wise adaptive step size $\boldsymbol{\mu}$ (Algorithm \ref{alg:me-bsb}, steps 15–17; Algorithm \ref{alg:se-sg-dsb}, steps 14–17).

When search is making consistent progress, the step size moderately increases with $R$ to accelerate exploration; as the population gradually freezes, the step size automatically decays with rising $F$ to stabilize refinement. The diversity-gated guidance (Eq.~\ref{eq:guidance}) resolves the population homogenization problem caused by GSB's constant global guidance~\cite{xiaoGloballyGuidedSimulated2026}: when diversity is high, elite guidance fully exploits high-quality solution information; when diversity declines into the warning zone, the gating factor automatically decays, weakening elite attraction and preserving residual search freedom.

Ablation experiments (\ref{app:ablation}) confirm that the exploration subpopulation is the primary pillar of population quality (its removal causes catastrophic degradation), the rescue mechanism ranks second, and density-aware adaptive scheduling is the key enabler of SG-DSB's generalization on heterogeneous benchmarks.

In terms of computational efficiency, the cost of each step in AE-QSB is primarily attributed to the SB dynamics updates, which include sparse matrix–dense matrix multiplication $J \cdot \boldsymbol{\phi}$ with complexity $O(\text{nnz} \cdot B)$, and amplitude momentum updates with complexity $O(NB)$. The perception layer is triggered once every $\Delta_{\text{eval}}$ steps and performs only statistical calculations $O(NB)$ of $(D, F, Q, R)$ based on the obtained $C_b$, without requiring additional matrix multiplications.
\section{Algorithm instances and experimental evaluation}
\label{sec:experiments}

\subsection{ME-BSB: $F$-driven adaptive switching}
\label{sec:me-bsb}
ME-BSB triggers the switch from BSB to DSB based on a freezing rate $F$ (Eq.~\ref{eq:mode}), combined with a minimum time gating condition $\tau \geq \tau_{\min}$, enabling automatic phase transition without manual intervention. The design concept can be summarized as "exploration first, commitment later": during the first half of the evolution, BSB continuous coupling is employed for breadth-first basin discovery, while in the second half, DSB discrete coupling is used for depth-first final state refinement. The continuous amplitude-preserving gradient information of BSB allows ME-BSB to rapidly locate high-quality basins in the early stages of evolution ($T \approx 100$) and maintain a leading position in the optimal individual dimension. However, after the subsequent abrupt switch to DSB, the momentum jump disrupts the continuity of refinement, resulting in a final average mass (Gap 0.26\%) that is lower than that achieved by a subsequent smooth transition scheme. Single-elite guidance ($K = 1$, Hamming-distance filter $\theta_H = 0.02$) with a 15\% exploration subpopulation ($\rho_{\text{explore}} = 0.15$) preserves guidance efficiency while avoiding premature population homogenization.

\begin{algorithm}[!t]
\footnotesize
\caption{ME-BSB: $F$-Driven Mode-Switching Enhanced BSB-to-DSB}
\label{alg:me-bsb}
\begin{algorithmic}[1]
\STATE Parameters: $\gamma=0.80$, $\mu_0=1.0$, $a(0)=1.0$, $F_{\text{switch}}=0.24$, $\beta_{\text{dense}}=0.08$, $\tau_{\min}=0.18$, $\rho_{\text{explore}}=0.15$, $\rho_{\text{explore}}^{\text{late}}=0.05$
\STATE \hspace{2em} $K=1$, $\alpha_{\text{gbest}}=0.16$, $\alpha_{\text{tabu}}=0.08$, $T_{\text{restart}}=300$, $T_{\text{stall}}=50$, $F_{\text{early}}=0.98$, $\tau_{\text{sprint}}=0.66$
\STATE $\mathbf{x},\mathbf{y} \sim \mathcal{U}(-0.1,0.1)^{N\times B}$, $\mathcal{E}=\varnothing$, $\texttt{in\_dsb}=\text{False}$
\STATE $\mathcal{M}_{\text{explore}} = \lfloor\rho_{\text{explore}}B\rfloor$ exploration columns (never accept elite guidance)
\FOR{$t = 0,\dots,T-1$, $\tau=t/T$}
    \IF{$\tau\geq\tau_{\text{sprint}}$}
        \STATE $\mathcal{M}_{\text{explore}}$ shrinks to $\lfloor\rho_{\text{explore}}^{\text{late}}B\rfloor$
    \ENDIF
    \IF{$t \bmod \Delta_{\text{eval}} = 0$}
        \STATE $\mathbf{s}=\operatorname{sgn}(\mathbf{x})$; compute cuts (Eq.~\ref{eq:cut-mat}); update per-column best cache
        \STATE Update elite pool $\mathcal{E}$ ($K=1$, $\theta_H=0.02$, Eq.~\ref{eq:ham-filter})
        \STATE Compute $D$ (Eq.~\ref{eq:D-def}), $F$ (Eq.~\ref{eq:F-def}), $Q$ (Eq.~\ref{eq:Q-def}), $R$ (Eq.~\ref{eq:R-def}); $\mu_b$ (Eq.~\ref{eq:mu-b})
        \IF{$\neg\texttt{in\_dsb} \land F > \min(0.95,\,F_{\text{switch}}+\beta_{\text{dense}}(1-\tau)) \land \tau\geq\tau_{\min}$}
            \STATE $\texttt{in\_dsb}=\text{True}$ \COMMENT{$F$-driven irreversible switch (Eq.~\ref{eq:mode})}
        \ENDIF
        \IF{$D<0.3$} \STATE greedy 1-flip \ENDIF
        \IF{$D<0.25 \land \tau<0.7$} \STATE emergency restart (30\% worst) \ENDIF
        \IF{$(t+1)\bmod T_{\text{restart}}=0$} \STATE elite restart (Eqs.~\ref{eq:restart}-\ref{eq:sigma}) \ENDIF
        \IF{$\tau\geq\tau_{\text{sprint}} \land (t+1)\bmod 160 = 0$}
            \STATE elite blend worst 10\% columns + bitflip top-10\% columns
        \ENDIF
        \IF{$\texttt{stall}>T_{\text{stall}} \land F>F_{\text{early}} \land Q<0.05$} \STATE \textbf{break} \ENDIF
        \STATE $\mathbf{g}_{\text{guide}} \gets$ elite sign pattern ($N\times 1$ broadcast to $B$ columns)
    \ENDIF
    \STATE \COMMENT{Execution layer (every step)}
    \STATE $a(t)=a(0)\tau^{\gamma}$; $\boldsymbol{\phi}=\mathbf{x}$ if $\neg\texttt{in\_dsb}$ else $\operatorname{sgn}(\mathbf{x})$ (hard switch)
    \STATE Compute $\mathbf{g}$ (Eq.~\ref{eq:guidance}, exploration columns set to zero)
    \STATE Update $\mathbf{y}, \mathbf{x}$ (Eq.~\ref{eq:sb-disc}, with $\boldsymbol{\phi},\mathbf{g},\boldsymbol{\mu}$)
\ENDFOR
\STATE \RETURN $\{\mathbf{s}_b,C(\mathbf{s}_b)\}_{b=1}^{B}$
\end{algorithmic}
\end{algorithm}

The core design philosophy of ME-BSB is to let the population state--rather than a preset iteration count--autonomously determine the evolutionary phase. The switching moment is dominated by the population's own freeze level rather than by time progression, hence the designation $F$-driven. In the early evolution, BSB continuous coupling preserves amplitude gradient information, facilitating global basin discovery; in the terminal evolution, DSB discrete coupling eliminates amplitude noise and acts directly on the Ising energy gradient. The exploration subpopulation (15\% of columns) receives no elite guidance whatsoever, preserving search freedom independent of the elite direction: on frustrated graphs, some high-quality basins lie at large Hamming distances from the elite basin, unreachable by local perturbations alone; the exploration subpopulation provides a channel for reaching remote basins through natural dynamical drift.

Initialization (steps 1–2) uses a small-range uniform distribution, an empty elite pool, and randomly selects $\rho_{\text{explore}}$ columns as the exploration subpopulation. During the sprint phase ($\tau \geq \tau_{\text{sprint}}$), the exploration subpopulation automatically shrinks to $\rho_{\text{explore}}^{\text{late}}$, releasing computation for elite refinement.

The perception layer (steps 5–8) calculates the population state at every $\Delta_{\text{eval}}$ step: it discretizes the amplitude to $\mathbf{s}$, evaluates the cut value (Eq.~\ref{eq:cut-mat}), updates the column-by-column optimal cache and elite pool ($K=1$, Eq.~\ref{eq:ham-filter}), calculates the four indicators $(D,F,Q,R)$, and determines the step size $\mu_b$ (Eq.~\ref{eq:mu-b}). The column-by-column optimal cache ensures that even if the exploration column drifts into a suboptimal region at a later stage, the final output is taken from the main peak and is not affected by tail outliers.

The coupling mode switching (step 9) is determined by Eq.~\ref{eq:mode}. The dynamic threshold $F_{\text{switch}} + \beta_{\text{dense}}(1-\tau)$ leverages the complementarity of time and freezing: early on, when $(1-\tau)$ is large, the threshold is raised to suppress premature discretization; in the late stage, as $(1-\tau)$ decays to zero, the threshold drops to $F_{\text{switch}}$. The parameter $\tau_{\min}$ sets a minimum time limit to prevent initial fluctuations during the startup phase from triggering false switching. Once triggered, the switch is irreversible: $\phi = \mathbf{x}$ permanently switches to $\phi = \operatorname{sgn}(\mathbf{x})$.

Three protection mechanisms (steps 10–12) address risks at different stages. Greedy 1-flip ($D < 0.3$) injects local perturbation at extremely low overhead by flipping only the single most gainful bit per column. Elite restart (Eqs.~\ref{eq:restart}-\ref{eq:sigma}) periodically replaces the worst columns with elite plus noise; noise decays with $\tau$, and the tabu direction enforces exploration of novel regions. During the sprint phase, at fixed intervals, elite blending is applied to the worst 10\% columns (replacing a fraction of the column state with the corresponding elite component, blending ratio $\approx 0.7$), while multi-bit deep refinement is applied to the top-10\% columns. Early stopping terminates when improvement stagnation, amplitude saturation, and sign locking are simultaneously satisfied.

The execution layer updates every step: the guidance vector $\mathbf{g}$ is computed via Eq.~\ref{eq:guidance} but set to zero for exploration columns; the bifurcation parameter follows $a(t) = a(0)\tau^\gamma$; the coupling function hard-switches based on the $\texttt{in\_dsb}$ flag; amplitudes and momenta are updated per Eq.~\ref{eq:sb-disc}.

The three core design decisions underlying the above process distinguish ME-BSB from similar methods. First, the function $F$ drives hard switching, differing from the traditional approach of dividing by iteration count or time. ME-BSB determines when to shift from exploration to refinement based on the population's freezing degree, naturally adapting to the varying convergence speeds of different graph structures. Second, the diversity-gated self-stabilizing loop, $\min(D/D_{\text{thresh}}, 1)$, creates negative feedback between elite guidance intensity and population diversity. Elite guidance is further regulated by stage scheduling $\omega(\tau)$ , producing a synergistic effect characterized by "early preloading—mid-term self-stabilization—sprint amplification." Third, 15\% of the columns do not accept guidance at all; these are called weak probing subgroups, which preserve the population's independent random exploration capability.

\subsection{SE-DSB / SG-DSB: Smooth exploration and density-aware enhancement}
\label{sec:se-sg-dsb}
SE-DSB replaces hard switching with linear mixed coupling defined as $\phi = (1-r_t)\mathbf{x} + r_t \operatorname{sgn}(\mathbf{x})$ (Eq.~\ref{eq:mixed-coupling}) to eliminate the risk of critical point momentum jumps. The SE-DSB evolution process retains both continuous exploration and discrete refinement components, embodying the design principle of "exploring while refining." The parameter $r_0=0.48$ represents a mixed evolutionary coupling of 52\% BSB and 48\% DSB. The parameter $r_t$ increases smoothly with $\tau$ and $F$ (Eq.~\ref{eq:r-schedule}), allowing the weights of the continuous and discrete components to transition gradually without requiring explicit stage partitioning. The core advantage of linear mixing is the elimination of momentum jumps caused by hard switching, which is a refinement rather than a post-remedy. 

SG-DSB introduces two enhancements based on SE-DSB. The density-aware adaptive parameter scheduling of SG-DSB (Eq.~\ref{eq:density-s}) dynamically adjusts eight parameters, including guidance strength, gate control lower limit, and switching threshold, enabling the algorithm to switch adaptively. SG-DSB is also inspired by GSB for velocity momentum (Equation 20) and applies an exponential moving average (EMA) to the elite guidance direction. Both methods share the core architecture of single elite guidance ($K = 1$), Hamming distance filtering ($\theta_H = 0.02$), an 18\% exploration subgroup ($\rho_{\text{explore}} = 0.18$), gated rescue (Eq.~\ref{eq:rescue}), and Tabu restart (Eq.~\ref{eq:restart}). The complete process is detailed in Algorithm \ref{alg:se-sg-dsb}.

\begin{algorithm}[!t]
\footnotesize
\caption{SE-DSB / SG-DSB: Smooth Exploration And Density-aware Enhancement}
\label{alg:se-sg-dsb}
\begin{algorithmic}[1]
\STATE Parameters: $\gamma=0.80$, $\mu_0=1.0$, $\mu_{\min}=0.40$, $\mu_{\max}=1.80$, $\rho_R=0.40$, $\rho_F=0.70$, $\alpha_{\text{gap}}=0.30$
\STATE \hspace{2em} $r_0=0.48$, $\kappa_{\tau}=0.66$, $\kappa_F=0.20$, $F_{\text{switch}}=0.23$, $\tau_{\text{fallback}}=0.44$
\STATE \hspace{2em} $\rho_{\text{explore}}=0.18$, $\rho_{\text{explore}}^{\text{late}}=0.06$, $\tau_{\text{sprint}}=0.64$, $K=1$, $\theta_H=0.02$, $\alpha_{\text{tabu}}=0.10$
\STATE \hspace{2em} $\tau_{\text{resc}}=0.34$, $\lambda_{\text{rescue}}=0.78$, $T_{\text{stall}}=50$, $F_{\text{early}}=0.98$
\STATE \hspace{2em} \textbf{SG-DSB only:} $\alpha_{\text{init}}=0.90$, $\alpha_{\text{end}}=0.45$, $b_{\text{cos}}=0.33$, $p_{\text{decay}}=0.55$
\STATE Proportional-sign initialization: $x_{i,b}=\pm0.80(0.8+0.4\,\mathcal{U})$ w.p.\ $p\in\{0.2,0.35,0.5\}[b\bmod 3]$
\STATE $\mathbf{y}=\mathbf{0},\;\mathcal{E}=\varnothing,\;\texttt{ph2}=\text{False},\;\mathbf{m}=\mathbf{0}$; $\mathcal{M}_{\text{explore}} \gets$ randomly select $\lfloor B\rho_{\text{explore}}\rfloor$ columns
\FOR{$t = 0,\dots,T-1$, $\tau=t/T$}
    \IF{$\tau\geq\tau_{\text{sprint}}$}
        \STATE $\mathcal{M}_{\text{explore}}$ shrinks to $\lfloor B\rho_{\text{explore}}^{\text{late}}\rfloor$
    \ENDIF
    \IF{$t \bmod \Delta_{\text{eval}} = 0$}
        \STATE $\mathbf{s}=\operatorname{sgn}(\mathbf{x})$, cuts (Eq.~\ref{eq:cut-mat}); update per-column best cache; update $\mathcal{E}$ (Eq.~\ref{eq:ham-filter})
        \STATE Compute $D$ (Eq.~\ref{eq:D-def}), $F$ (Eq.~\ref{eq:F-def}), $Q$ (Eq.~\ref{eq:Q-def}), $R$ (Eq.~\ref{eq:R-def}); $\mu_b$ (Eq.~\ref{eq:mu-b})
        \IF{$D<0.3$} \STATE greedy 1-flip \ENDIF
        \IF{$D<0.25 \land \tau<0.7$} \STATE emergency restart (30\%) \ENDIF
        \IF{$(t+1)\bmod 280 = 0$} \STATE Tabu restart (Eq.~\ref{eq:restart}) \ENDIF
        \IF{$\tau>0.3$} \STATE bitflip top-15\% (Eq.~\ref{eq:bitflip}) \ENDIF
        \IF{$\tau\geq\tau_{\text{resc}}$} \STATE gated rescue (Eq.~\ref{eq:rescue}) \ENDIF
        \IF{$\tau\geq\tau_{\text{sprint}}$} \STATE bitflip top-12\% \ENDIF
        \IF{$\texttt{stall}>T_{\text{stall}} \land F>F_{\text{early}} \land Q<0.05$} \STATE \textbf{break} \ENDIF
        \STATE $\mathbf{g}_{\text{guide}} \gets$ elite sign pattern (broadcast to $B$ columns)
        \IF{SG-DSB}
            \STATE velocity momentum EMA update + cosine check (Eq.~\ref{eq:momentum})
        \ENDIF
    \ENDIF
    \STATE \COMMENT{Execution layer (every step)}
    \STATE $a(t)=a(0)\tau^{\gamma}$
    \IF{$\neg\texttt{ph2} \land (F>F_{\text{switch}} \lor \tau>\tau_{\text{fallback}})$}
        \STATE $\texttt{ph2}=\text{True}$
    \ENDIF
    \STATE Compute $\boldsymbol{\phi}$ (Eqs.~\ref{eq:mixed-coupling}-\ref{eq:r-schedule}); $\mathbf{d}_{\text{guide}}=\mathbf{m}$ (SG-DSB) or $\mathbf{g}_{\text{guide}}$ (SE-DSB)
    \STATE Compute $\mathbf{g}$ (Eq.~\ref{eq:guidance}, exploration columns receive weak guidance $0.30s\cdot w_{\text{eff}}$)
    \STATE Update $\mathbf{y}, \mathbf{x}$ (Eq.~\ref{eq:sb-disc}, with $\boldsymbol{\phi},\mathbf{g},\boldsymbol{\mu}$); if out-of-bounds, $y_{i,b}=0$
\ENDFOR
\STATE \RETURN $\{\mathbf{s}_b,C(\mathbf{s}_b)\}_{b=1}^{B}$
\end{algorithmic}
\end{algorithm}

The design of SE-DSB/SG-DSB primarily aims to achieve a smooth convergence transition while maintaining exploration capability. The linear mixed coupling $\phi$ transforms the hard switching of ME-BSB into a continuous process, thereby eliminating the risk of momentum jumps at critical points. The core of density perception in SG-DSB is the logarithmic scale of graph density, defined as $d = |E| / [N(N-1)/2]$.
\begin{equation}
s = \operatorname{clip}\!\left(\frac{\log(d/0.005)}{\log(1/0.005)},\;0,\;1\right) \in [0,1],
\label{eq:density-s}
\end{equation}
Among these, $s=0$ corresponds to extremely sparse graphs, while $s=1$ corresponds to complete graphs. The parameter $s$ is used to dynamically adjust variables such as guidance strength, gating lower limit, and switching threshold, enabling the algorithm to adaptively transition between sparse and dense graphs. This adaptive capability is the primary advantage that distinguishes SG-DSB from SE-DSB.

Initialization (steps 1–2) employs a proportional sign strategy: column $B$ is assigned alternately with three positive sign probabilities of $\{0.2, 0.35, 0.5\}$, ensuring that the initial population spans a wide symbol space ranging from negative to positive values. With zero momentum, SG-DSB maintains an additional vector $\mathbf{m}$. Steps 3–7 correspond to the relevant procedures described in the ME-BSB method above. The greedy 1-flip, emergency restart (step 8), and Tabu elite restart (step 9) within the protection mechanism function identically to those in the ME-BSB algorithm.

The coupling mode (steps 14–15) is the core mechanism that distinguishes SE-DSB from ME-BSB, employing a two-stage smooth transition strategy.
\begin{equation}
\phi(\mathbf{x}_b; r_t) = (1 - r_t) \cdot \mathbf{x}_b + r_t \cdot \operatorname{sgn}(\mathbf{x}_b),
\label{eq:mixed-coupling}
\end{equation}
\begin{equation}
r_t = \begin{cases}
\operatorname{clip}(r_0 + \kappa_\tau \tau + \kappa_F F,\; 0,\; 1), & \text{phase 1}, \\
r_{\text{target}} + (1 - r_{\text{target}}) \cdot \frac{t - t_{\text{switch}}}{\Delta_{\text{ramp}}}, & \text{phase 2}.
\end{cases}
\label{eq:r-schedule}
\end{equation}
In Eq \ref{eq:r-schedule}, the variable $r_t$ in Phase 1 monotonically increases with normalization time $\tau$ and freezing rate $F$. The initial value $r_0 \approx 0.48$ assigns nearly equal weights to the continuous and discrete components. Phase 2 is triggered either when $F > F_{\text{switch}}$ or $\tau > \tau_{\text{fallback}}$, and $r_t$ smoothly rises to pure DSB within $\Delta_{\text{ramp}}$ steps. These two triggering conditions are complementary: the former ensures that the population is fully frozen, while the latter provides a fallback when freezing is slow.

SG-DSB exclusive speed momentum (step 13) applies EMA smoothing to the elite guidance direction.
\begin{equation}
\begin{aligned}
\mathbf{m}^{(t+1)} &= \alpha_{\text{mom}} \cdot \mathbf{m}^{(t)} + (1 - \alpha_{\text{mom}}) \cdot \mathbf{g}_{\text{guide}}, \\
\alpha_{\text{mom}}(\tau) &= \alpha_{\text{init}} - (\alpha_{\text{init}} - \alpha_{\text{end}}) \cdot \tau^{\,p_{\text{decay}}},
\end{aligned}
\label{eq:momentum}
\end{equation}
Early high $\alpha_{\text{mom}}$ allows the momentum direction $\mathbf{m}^{(t)}$ to play a greater role, suppressing the interference of single-step guided fluctuations on the search trajectory. In contrast, late low $\alpha_{\text{mom}}$ causes the momentum $\mathbf{g}_{\text{guide}}$ to quickly guide the direction, which is suitable for high-precision directional requirements during the refinement stage. The cosine synergy check prevents conflicts between the smooth direction and the momentum $\mathbf{y}$. SE-DSB does not include this mechanism and directly uses the instantaneous value of $\mathbf{g}_{\text{guide}}$ as the guiding direction.

Gated rescue (Step 10) is activated when the gating conditions $\tau \geq \tau_{\text{resc}}$ are met, and elite worst re-anchoring is performed on the worst column.
\begin{equation}
\mathbf{x}_{\text{worst}} = \lambda_{\text{rescue}} \cdot \mathbf{x}_{\text{elite}} + (1 - \lambda_{\text{rescue}}) \cdot \mathbf{x}_{\text{worst}} + \sigma(\tau) \cdot \mathcal{N}(0, I),
\label{eq:rescue}
\end{equation}
Retain some of the original state of $1-\lambda_{\text{rescue}}$  to prevent the rescue column from completely losing its original search information.

The launch timing of the Bit-flip refinement (steps 10–11) occurs earlier than that of ME-BSB. If $\tau > 0.3$, adaptive flipping is applied to the top 15\% of columns, and the sprint period is upgraded to the top 12\%. The number of flips performed each time is dynamically determined by the maturity factor $F \cdot (1-Q)$. This approach initiates moderate refinement of already stable columns in the middle of the evolution process, rather than waiting for a global sprint.
\begin{equation}
n_{\text{flip}} = n_{\min} + (n_{\max} - n_{\min}) \cdot F \cdot (1 - Q).
\label{eq:bitflip}
\end{equation}
The ME-BSB exploration column is entirely unguided, whereas the SE-DSB/SG-DSB exploration column incorporates density-aware weak guidance with an intensity of $0.30s \cdot w_{\text{eff}}$. The lower limit of diversity gating is also modulated by $s$, with sparse graphs maintaining higher lower limits to ensure a minimum guidance direction.

The execution layer (steps 14–17) updates the amplitude and momentum according to Eq. (\ref{eq:sb-cont}) and clears any out-of-bounds momentum. At the end of the evolution, an adaptive bit-flip is applied to the rescue column, and the column-by-column optimal cache, similar to ME-BSB, ensures that the output is not affected by tail outliers.

\subsection{Experimental setup}
\label{sec:setup}

The experimental evaluation is conducted at two levels. The first level takes G22 ($N = 2000$, frustrated graph, optimum 13359, $\approx 2\times 10^4$ edges) as the representative instance for depth analysis with $T = 1000$, while also scanning $T = 10$-$1000$ for convergence behavior analysis. The second level evaluates generalization on 71 heterogeneous G1-G81 instances ($N = 800$-$20000$).

The eight compared algorithms are organized into four groups: (1) Standard (Standard BSB / Standard DSB, basic SB dynamics); (2) AE-QSB (ME-BSB / SE-DSB / SG-DSB); (3) GSB~\cite{xiaoGloballyGuidedSimulated2026} (GSB-BSB / GSB-DSB, global guidance); (4) Tabu-SB~\cite{taoTabuenhancedSimulatedBifurcation2026} (Tabu-BSB / Tabu-DSB, tabu enhancement). AE-QSB variants retain only per-column best-cache recovery and lightweight adaptive bitflip; no additional postprocessing pipeline is introduced.
 
All algorithms are implemented based on a unified base class interface (with batch size $B = 256$, float32 precision, PyTorch 2.x + CUDA, NVIDIA RTX GPU). Sparse matrices are stored in CSR (Compressed Sparse Row) format. The complete code is planned to be made publicly available after the paper is published~\cite{liuAEQSBSourceCode2026}. The ablation variant selectively enables or disables the corresponding mechanisms within the same dynamic framework by controlling flags. The G22 experiment independently repeats each algorithm and iteration configuration 10 times (using deterministic seeds), reporting the mean, standard deviation, and quantiles across repetitions. Among these, SE-DSB and SG-DSB employ a multi-start strategy (for $T \geq 250$, independent starts are repeated 3–5 times; candidate solutions from the top two starts are selected and evaluated according to the (mean, max) lexicographic order), while other algorithms perform a single start per run. Algorithm comparisons use a two-sided Welch’s $t$-test with $\alpha = 0.05$.

Ablation experiments adopt a two-tier design (details in \ref{app:ablation}); G22 ablation is evaluated at $T = 1000$, $B = 256$, with 5 independent repeats. G1-G81 experiments also use 5 repeats per graph, with adaptive iteration budgets per graph category.

\subsection{Depth analysis on G22}
\label{sec:g22}

Under $T=1000$, $B=256$ (10 independent repeats), SE-DSB and SG-DSB jointly achieve the best mean quality, with gaps both below 0.05\%; the difference between them lies within statistical noise. The two alternate in leading throughout the convergence process: SG-DSB holds a slight advantage during $T=250$-$500$, SE-DSB overtakes after $T=750$, and the terminal performance is nearly identical, suggesting that under the current parameter configuration their difference primarily stems from random fluctuations in multi-launch sampling rather than systematic architectural superiority. ME-BSB, with a single launch, approaches the extreme-value performance of the former two, and achieves the optimal time-quality trade-off, attaining mean quality far superior to the Standard baselines with significantly shorter runtime. Complete performance data are summarized in Table~\ref{tab:summary}.

Standard baselines reveal the limitations of fixed scheduling from the negative perspective. Standard BSB plateaus early in the evolution ($T \approx 100$) with virtually no further improvement; all repeats systematically converge to the same suboptimal basin. Standard DSB can intermittently discover high-quality individual solutions, but exhibits enormous cross-repeat variance and mean quality far inferior to AE-QSB methods, indicating that discrete coupling under fixed scheduling lacks mechanisms to sustain population quality. GSB-BSB goes to the opposite extreme: within a single repeat the population nearly completely collapses (intra-repeat variance approaching zero), and all repeats collapse near the same suboptimal basin; the cross-repeat consistency paradoxically masks its systematically suboptimal nature, as the lack of a diversity self-stabilization mechanism deprives the algorithm of the ability to escape the first-encountered basin under strong guidance.

\begin{figure}[!t]
\centering
\aeqsbincludegraphics[width=0.95\textwidth]{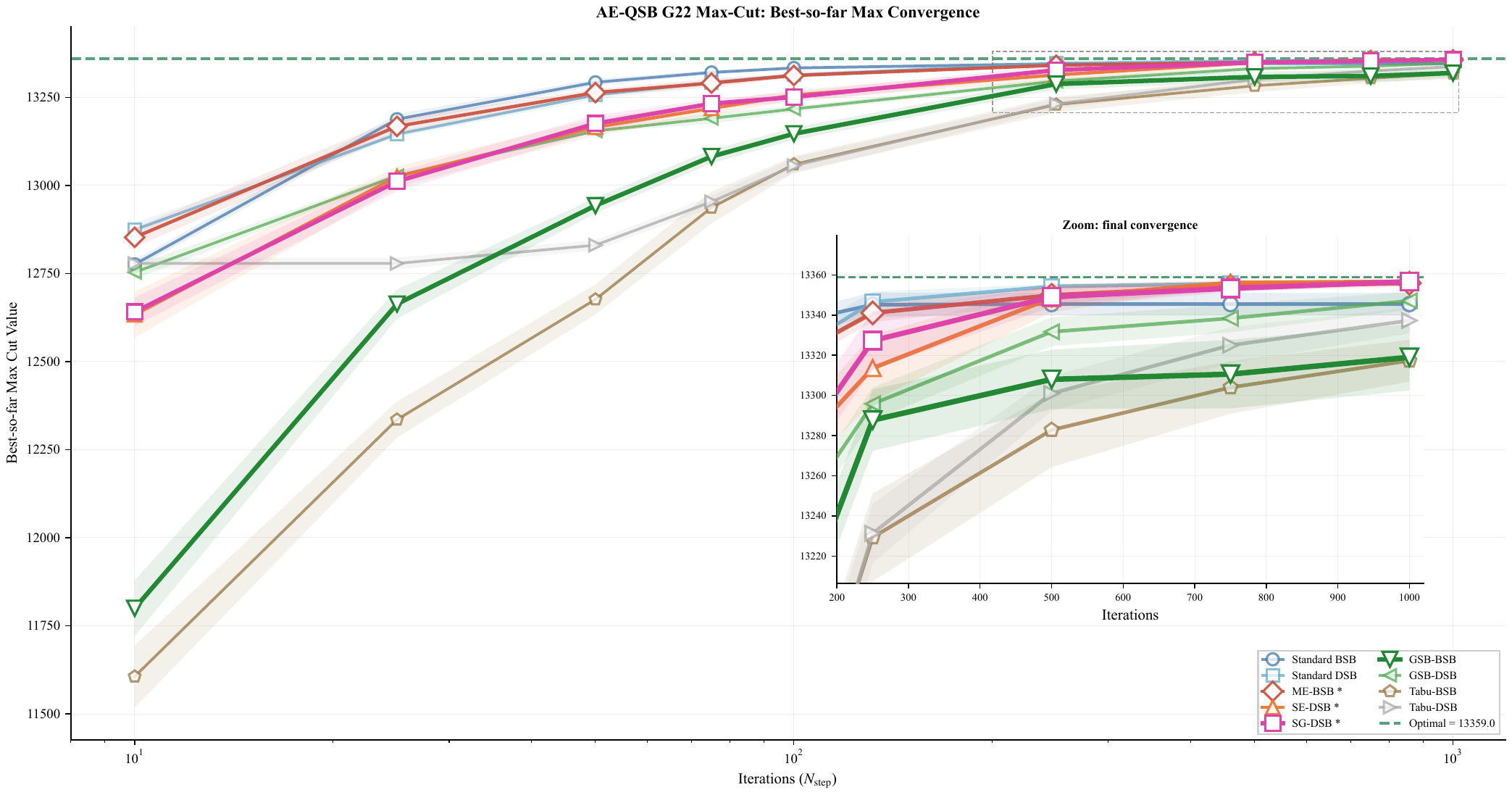}
\caption{G22 best-value convergence curves. Main panel: global-best cut values across $T = 10$-$1000$ (solid lines = mean of 10 repeats, shading = $\pm 1$ std, dashed line = known optimum 13359). Inset: zoom of terminal convergence region (iterations 200-1100).}
\label{fig:g22-conv}
\end{figure}

\begin{figure}[!t]
\centering
\aeqsbincludegraphics[width=0.95\textwidth]{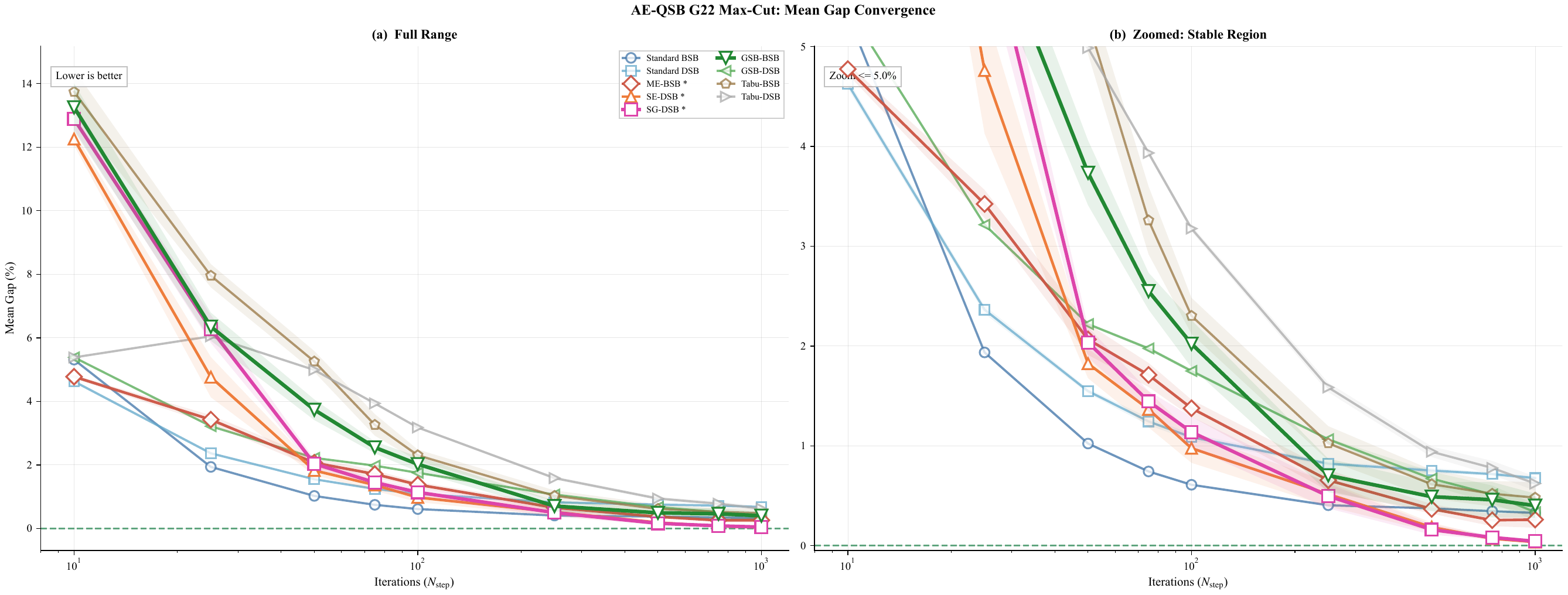}
\caption{G22 mean gap convergence curves. Left: complete gap convergence. Right: stabilized-region zoom (y-axis ceiling adaptive to 95th percentile of gap $\leq$ 5\% data).}
\label{fig:g22-gap}
\end{figure}

From the convergence trajectories, the three AE-QSB instances exhibit characteristic evolutionary patterns. ME-BSB benefits from the early exploration efficiency of BSB continuous coupling, leading in the extreme-value dimension from the very early evolution, with the improvement rate gradually slowing as iteration progresses, displaying a smooth convergence morphology. SG-DSB and SE-DSB exhibit a catch-up-and-converge trajectory: lagging behind most algorithms early, accelerating improvement in mid-evolution ($T = 250$-$500$), temporally aligned with rescue activation ($\tau \geq 0.34$) and maturity-gated bitflip onset ($\tau > 0.30$), and jointly converging to near-optimum levels at termination. Figure \ref{fig:g22-conv} shows the evolution curves of the optimal cut values for each algorithm over the number of iterations, while Figure \ref{fig:g22-gap} illustrates the convergence process of the mean gap.

\begin{figure}[!t]
\centering
\aeqsbincludegraphics[width=0.90\textwidth]{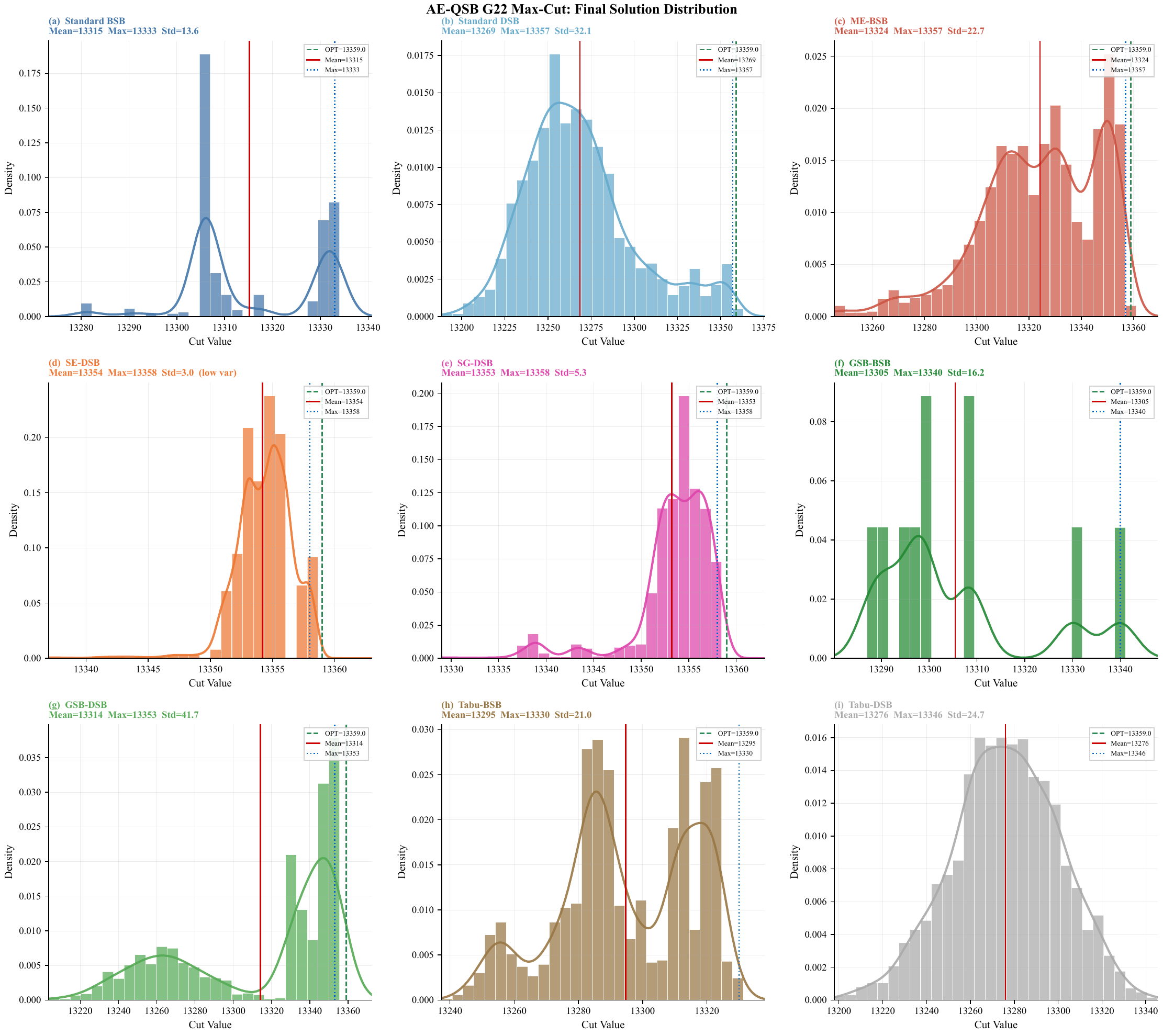}
\caption{G22 final solution distributions. Histograms and KDE curves of cut values for $B=256$ candidates at $T=1000$ (10 repeats aggregated). ME-BSB: unimodal left-skewed; SG-DSB/SE-DSB: extremely narrow unimodal; GSB-BSB: discrete multi-modal.}
\label{fig:g22-dist}
\end{figure}

The final distribution of population cut values reveals three qualitatively distinct patterns. ME-BSB exhibits a unimodal, skewed distribution: the main peak is concentrated in the high-quality interval near the optimal value, while a longer left tail extends toward lower values. This indicates that most candidate solutions have stably converged to the high-quality basin, although a few columns remain trapped in far suboptimal regions during the exploration process. The optimal buffer, applied column by column, ensures that the final output is derived from the main peak and is not influenced by the left-tail outliers. SG-DSB and SE-DSB both display extremely narrow unimodal distributions, with populations highly concentrated within tight bands close to the optimum. These results reflect the synergistic effect of linear mixed coupling, smooth transition, and gated rescue mechanisms. Additionally, the multi-start strategy further reduces variance across repeated experiments. GSB-BSB exhibits a discrete multimodal distribution: within a single replicate, the population collapses to nearly a single configuration (with internal variance approaching zero), but different replicate experiments converge to different non-optimal basins. Tabu-BSB shows a narrow but suboptimal distribution, indicating that the algorithm became locked away from the optimal basin during the early stages of evolution. Figure \ref{fig:g22-dist} presents the distribution of cut values for the final population of each algorithm using histograms and Kernel Density Estimation (KDE) curves.

\begin{figure}[!t]
\centering
\aeqsbincludegraphics[width=0.95\textwidth]{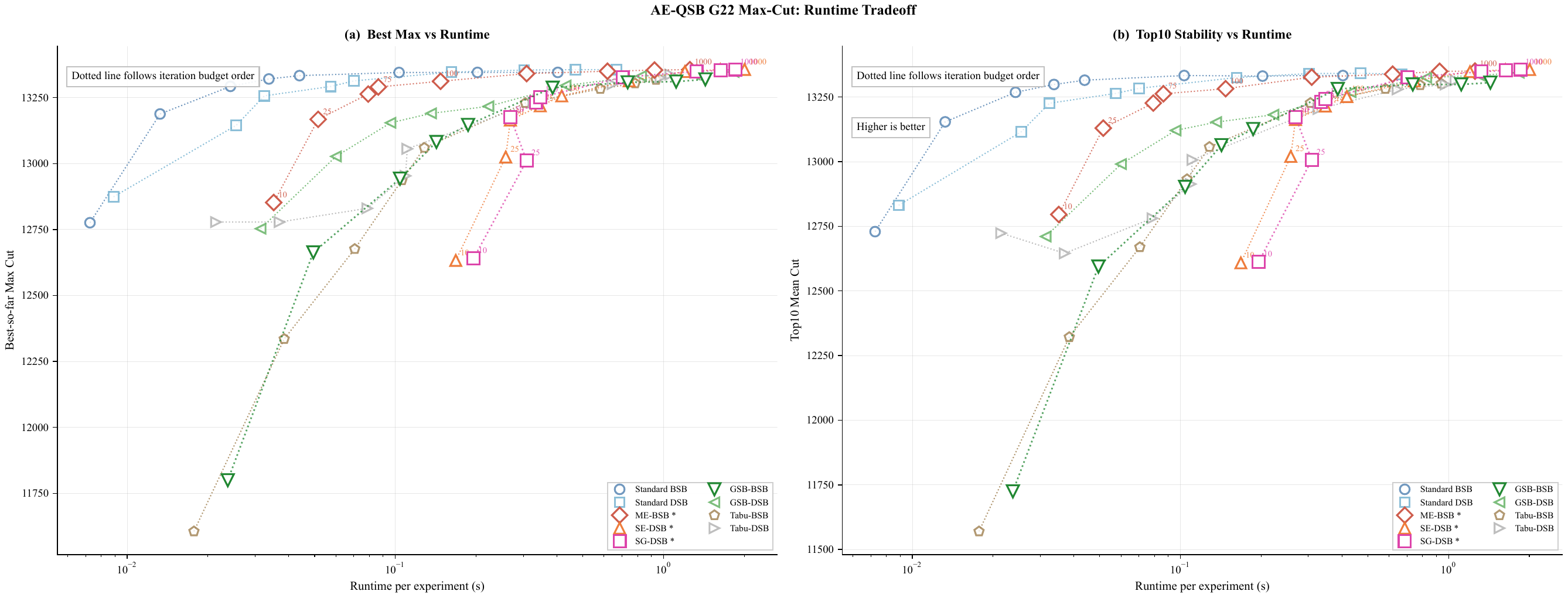}
\caption{G22 runtime-quality trade-off. Left: best cut vs.\ runtime scatter. Right: mean cut vs.\ runtime scatter. Each point represents one independent repeat; large markers indicate algorithm means. AE-QSB variants (diamonds) cluster toward the upper-left (high quality + low time).}
\label{fig:g22-time}
\end{figure}

In addition to the final solution quality, the computational efficiency of the algorithm is a crucial factor for practical deployment. Figure \ref{fig:g22-time} presents the running time and solution quality of each algorithm simultaneously in a scatter plot, clearly illustrating the trade-off patterns between efficiency and quality across different algorithms. ME-BSB achieves near-optimal extreme performance with the lowest single-run time, representing the best balance between time efficiency and solution quality at a single point. SE-DSB and SG-DSB trade off total running time across multiple startup strategies to enhance population-level reliability and temporal stability. The running time of the GSB variant is comparable to that of SE-DSB and SG-DSB; however, its solution quality is significantly lower, highlighting the efficiency cost associated with a consistently strong guidance strategy.

\begin{figure}[!t]
\centering
\aeqsbincludegraphics[width=0.90\textwidth]{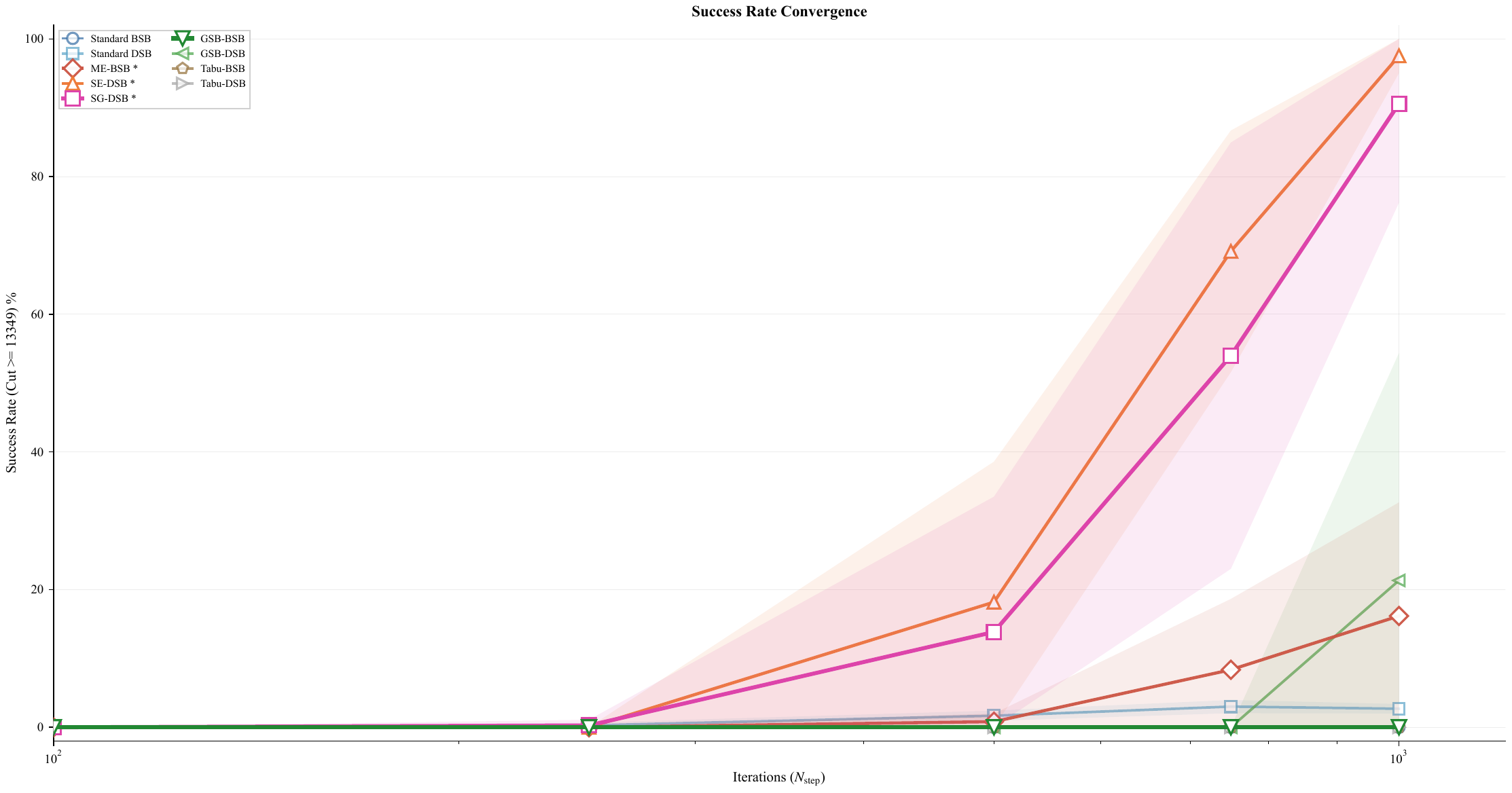}
\caption{G22 success-rate convergence. $P(\text{cut} \geq C_{\text{thresh}})$ vs.\ $T$ (10-repeat statistics), with the success threshold set near the known optimum. SE-DSB and SG-DSB both exceed 90\% terminal success rate under multi-launch.}
\label{fig:g22-success}
\end{figure}

The convergence curve of the success rate complements the conclusions drawn from the mean quality analysis. In Figure \ref{fig:g22-success}, the success rate of each algorithm on G22 is shown as a function of the number of iterations. As the number of iterations increases, the success rates of SE-DSB and SG-DSB steadily improve, both reaching high levels. Notably, SE-DSB achieves a higher success rate than SG-DSB across multiple bootstrapping strategies. The success rate of ME-BSB increases relatively steadily, consistent with its design strategy; further details can be found in \secref{sec:me-bsb}. Although a single run of ME-BSB may not always reach the optimal basin, it remains competitive at extreme values through column-by-column optimal caching. The success rates of Standard BSB and GSB-BSB remain at zero throughout, which aligns perfectly with the collapse behavior described earlier.
 
\subsection{G1-G81 heterogeneous benchmark}
\label{sec:g1g81}
\begin{figure}[!t]
\centering
\aeqsbincludegraphics[width=0.95\textwidth]{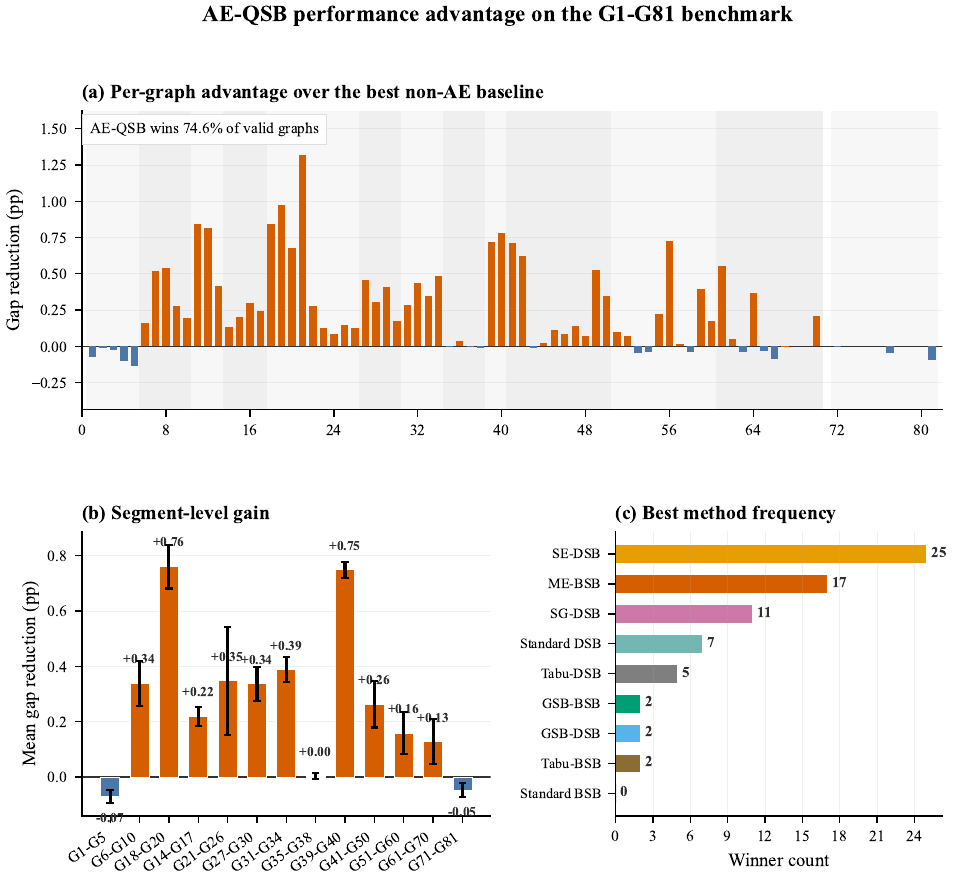}
\caption{G1--G81 comprehensive performance advantages. (a) Gap reduction per graph, with orange representing the optimal AE-QSB variant and blue representing the optimal non-AE baseline; the gray background indicates graph segmentation. (b) Average gap reduction for each segment, with numbers indicating AE-QSB's winning rate within that segment. (c) Algorithms sorted by Gap\_Mean across 71 graphs, showing how often each algorithm is the best method.}
\label{fig:g1g81}
\end{figure}

On 71 G-set instances, the execution strategy described in \secref{sec:setup} was employed. The statistical analysis of the optimal bandgap (Gap-Mean) obtained by each algorithm is presented in Table~\ref{tab:gset-freq}. Based on comprehensive performance advantages, the decomposition results are illustrated in Figure \ref{fig:g1g81}. The three variants of AE-QSB achieved optimal results on approximately three-quarters of the graphs; however, further improvement is needed for a small number of extremely sparse and dense graphs. The SE-DSB algorithm outperformed other algorithms in terms of the maximum number of optimal graphs, demonstrating the accuracy advantage of linear mixed coupling on near-optimal graphs (G21–G26, full coverage) and truly hard dense graphs (G27–G30). ME-BSB closely followed and performed exceptionally well with a weak guidance exploration strategy on some medium-scale dense maps (G41–G50, 6/10). SG-DSB ranked third, and in another dense subsegment (G31–G34, G39–G40, full coverage), the synergistic effect of density-aware adaptive scheduling and the rescue mechanism fully demonstrated the algorithm’s advantages. Standard DSB, GSB, and Tabu variants achieved optimal results on a few graphs, while the standard BSB did not achieve optimal results on any graph.

\begin{table}[!t]
\centering
\caption{Optimal frequencies for algorithms G1 to G81. The Gap-Mean data from multiple algorithms in the graph indicate that the algorithm with the lowest gap value performs best.}
\label{tab:gset-freq}
\footnotesize
\begin{tabularx}{\textwidth}{lccY}
\toprule
Algorithm & \# Optimal & Share (\%) & Typical dominant graph types \\
\midrule
\textbf{SE-DSB} & \textbf{25} & \textbf{35.2} & $N\approx2000$ near-optimal, truly-hard dense, frustrated spin glass \\
\textbf{ME-BSB} & \textbf{17} & \textbf{23.9} & Medium-scale dense, sparse unfrustrated, mixed scale (G35-G38 full) \\
\textbf{SG-DSB} & \textbf{11} & \textbf{15.5} & Extremely-hard dense sub-segment (G31-G34, G39-G40 full) \\
Standard DSB & 7 & 9.9 & Partial medium-scale graphs \\
Tabu-DSB & 5 & 7.0 & Large scale ($N\approx 6000$-$20000$) \\
Tabu-BSB & 2 & 2.8 & Easy sparse (G1-G5) \\
GSB-BSB & 2 & 2.8 & Easy sparse (G1-G5) \\
GSB-DSB & 2 & 2.8 & Isolated instances \\
\bottomrule
\end{tabularx}
\end{table}

Table \ref{tab:gset-freq} summarizes the optimal number of graphs identified by each algorithm across 71 graphs, based on Gap\_mean statistics (where a smaller Gap\_mean value indicates better performance). Figure \ref{fig:g1g81} illustrates the comprehensive performance advantages of AE-QSB over non-AE baselines from three perspectives: individual graphs, segmented groups, and overall statistics.

The segmented analysis Figure \ref{fig:g1g81}(a–b) reveals the boundaries of the regular migration of three algorithm instances along the graph's difficulty spectrum. The easily sparse segments (G1–G5) are dominated by GSB and Tabu variants with full coverage. The convergence efficiency of constant strong guidance and tabu exclusion on simple, non-setback graphs surpasses that of adaptive exploration. The setback spin glass segment (G6–G10) then comes into play, where AE-QSB achieves a 100\% success rate. ME-BSB leads with full coverage (4/4) in the sparse no-setback segment (G14–G17). The symbol mixing segment $\pm 1$ (G11–G13, G18–G20) is dominated by SE-DSB and SG-DSB with a 6/6 ratio, and the accuracy advantage of linear mixing coupling begins to emerge on complex symbol weight distributions. The near-optimal segment (G21–G26) and the truly hard dense segment (G27–G30) continue to be won by our AE-QSB framework, with a particularly significant gap reduction in the near-optimal segment.

In the dense subsegments G31–G34 and G39–G40, SG-DSB dominates with a score of 5/6, and the synergy between density-aware adaptive scheduling and rescue mechanisms is fully realized at extremely high densities. Standard DSB in G35–G38 reverses the trend with a score of 3/4, indicating that discrete coupling remains highly competitive in moderately complex energy landscapes. ME-BSB takes an overwhelming lead with a score of 9/10 in G41–G50; its weak guidance architecture proves more robust against deviations in reference values. The mixed-scale range (G51–G60) was statistically analyzed by AE-QSB, achieving a score of 7/10. In the large-scale graph (G61–G70), AE-QSB maintains a majority with a score of 5/8, while Tabu-DSB begins to demonstrate competitiveness with 2/8. Finally, Tabu-DSB dominates the ultra-large-scale segment (G71–G81) with full coverage (3/3), as its directional escape capability through taboo exclusion in large-scale sparse spaces surpasses adaptive exploration.

Examining the entire sequence, the winning range begins with Tabu/GSB, continues through AE-QSB, and ultimately reaches Tabu-DSB, forming a complementary chain that closely aligns with the graph's difficulty spectrum.

\subsection{Time-to-solution and approximation ratio analysis}
\label{sec:ttsar}
\begin{figure}[!t]
\centering
\aeqsbincludegraphics[width=0.90\textwidth]{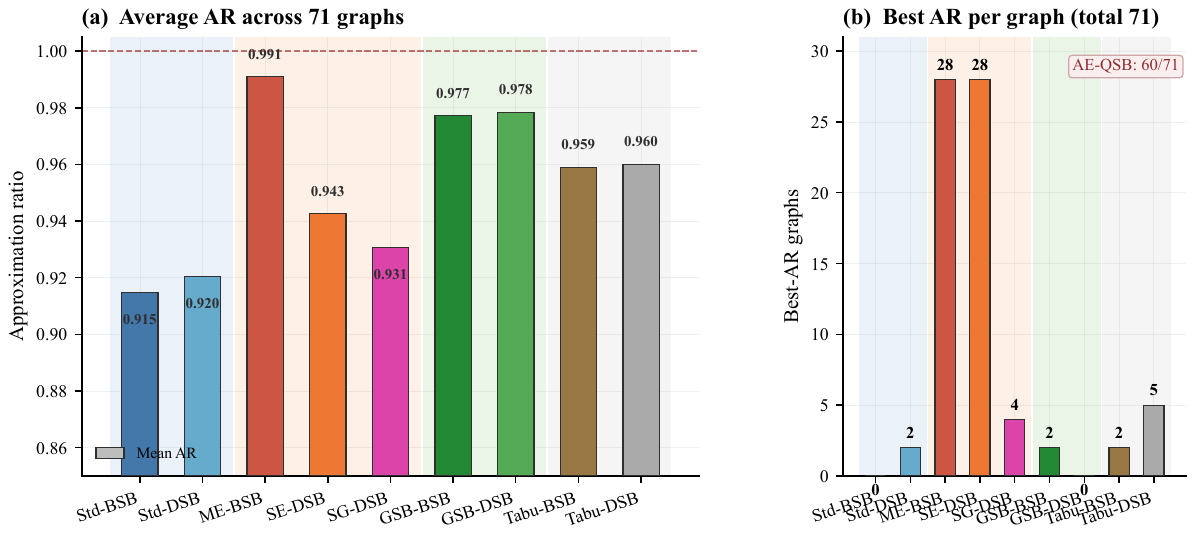}
\caption{Summary of approximate comparisons for G1–G81. (a) Bar chart showing the mean AR of each algorithm across 71 graphs, with a red dashed line at 1.0 and specific values indicated above each bar. (b) Number of graphs for which each algorithm achieves the optimal AR\_mean metric, along with their corresponding values. ME-BSB leads with a mean AR of 0.991 and the highest number of AR-best graphs.} 
\label{fig:ar-summary}
\end{figure}

In addition to solution quality, time efficiency and Approximation Ratio(AR) guarantees are also key factors in evaluating the practicality of algorithms. In this section, nine algorithms were systematically evaluated for their AR and Time to Solution (TTS) on all 71 graphs from G1 to G81. The AR is defined as $\text{AR} = C_{\text{alg}} / C_{\text{opt}}$, where $C_{\text{opt}}$ refers to the known best and proven optimal values. TTS measures the expected running time required for an algorithm to find the optimal solution with a probability not lower than the target at a given confidence level. If the single run time is $t_{\text{run}}$ and the success probability of a single experiment is $P_s$, then the TTS at a 99\% confidence level is given by $\text{TTS}_{0.99} = t_{\text{run}} \cdot \frac{\log(1 - 0.99)}{\log(1 - P_s)}$ Here, $P_s$ is estimated based on the empirical success frequency from 20 independent experiments for each instance. Under the multi-start strategy, $t_{\text{run}}$ represents the total time spent across multiple starts, and $P_s$ is the probability of reaching the optimal solution at least once. The AR and TTS statistics for each algorithm across the 71 graphs are summarized in Table~\ref{tab:ar-tts}, while the mean AR and the number of optimally solved graphs are shown in Figure \ref{fig:ar-summary}.

\begin{table}[!t]
\centering
\caption{Summary of solution time (TTS) and approximation ratio (AR) for graphs G1–G81. The median values are calculated from all 71 graphs. Bold indicates the optimal column. *Contains multiple startup strategies.} 
\label{tab:ar-tts}
\scriptsize
\begin{adjustbox}{max width=\textwidth}
\begin{tabular}{lcccc}
\toprule
Algorithm & AR Mean & AR-best graphs & TTS$_{0.99}$ median (s) & Median runtime (s) \\
\midrule
Standard BSB & 0.915 & 0 & 151.3 & 0.61 \\
Standard DSB & 0.920 & 2 & \textbf{2.3} & 0.70 \\
\textbf{ME-BSB} & \textbf{0.991} & \textbf{28} & 2.1 & 1.21 \\
SE-DSB & 0.943 & \textbf{28}$^*$ & 11.4$^*$ & 1.58 \\
SG-DSB & 0.931 & 4$^*$ & 13.0$^*$ & 1.62 \\
GSB-BSB & 0.977 & 2 & 43.1 & 1.81 \\
GSB-DSB & 0.978 & 0 & 53.5 & 1.85 \\
Tabu-BSB & 0.959 & 2 & 5877.5 & 1.20 \\
Tabu-DSB & 0.960 & 5 & 100.1 & 1.29 \\
\bottomrule
\end{tabular}
\end{adjustbox}
\end{table}
In the AR dimension, ME-BSB ranks first with a mean of 0.991, tied with SE-DSB for the highest number of AR-optimal images (28 each). The three variants of AE-QSB achieved AR optimality on approximately 80\% of the graphs. The AR mean of SE-DSB (0.943) was lowered by the ferromagnetic collapse on the no-setback graph, which resulted in a gap of 16\% to 30\% in the easily sparse segment (G1–G5), making the cross-graph mean appear very unfavorable. When these no-setback graphs are removed and the data reanalyzed, the median gap and ME-BSB are essentially the same (about 1.0\%), indicating that the accuracy of this architecture remains uncompromised in the graph classes where it excels. The AR mean of SG-DSB (0.931) is slightly lower than that of SE-DSB, reflecting a similar pattern observed on G22(\secref{sec:g22}): density-aware adaptive scheduling and velocity momentum provide gains on some graphs, but collapses on a few graphs offset these improvements. The average cross-graph AR for GSB-BSB (0.977) and GSB-DSB (0.978) is actually higher than that for SE-DSB and SG-DSB. However, the optimal number of AR graphs is only 2 and 0, respectively. This discrepancy between the mean and the optimal number of graphs may be due to the fact that constant strong guidance converges well on simple graphs, thereby increasing the mean, but performs poorly on the majority of more complex graphs.

In the TTS dimension, algorithms are generally categorized into three tiers. Standard DSB and ME-BSB belong to the first tier, with TTS medians slightly above two seconds (2.3s and 2.1s, respectively). However, ME-BSB's AR mean is nearly 7 percentage points higher. ME-BSB has significantly improved its quality without requiring much additional time. SE-DSB (11.4s) and SG-DSB (13.0s) spend more time on multiple startups in the second tier, effectively using several times the computational effort to compensate for cross-cycle reliability. Given that the gap pressure on G22 is below 0.05\%, this additional cost is justified. GSB and Tabu fall into the third tier. The TTS for the GSB variant reached 40–50 seconds, but the benefit of constant guidance was not realized; instead, time was wasted oscillating within an already locked suboptimal basin.

The median TTS of Tabu-BSB is 5,877.5 seconds, primarily due to the low value of $P_s$, where the denominator $\log(1-P_s)$ dominates the entire result as $P_s$ approaches zero in the TTS formula. Tabu-BSB failed to reach the optimal point in nearly half of the graphs after 20 trials, causing $P_s$ to be zero in those cases and dragging the median TTS to the hourly range. This value does not represent its typical running time; for the graphs it can solve (such as G1–G5), the TTS is only a few seconds. The value 5,877.5 seconds should be interpreted as an indicator that this algorithm did not perform well on almost half of the graphs. This also highlights the difficulty of using a single metric to accurately describe performance when the algorithm behaves very differently across instances. The TTS of Tabu-DSB (100.1 seconds) performs better overall, but there is still variation: it is strong on large-scale graphs but less effective on sparse setback graphs.

\section{Discussion}
\label{sec:discussion}
For the purpose of cross-level comparison, Table~\ref{tab:summary} summarizes the various indicators of each algorithm on G22 and G1–G81.

\begin{table}[!t]
\centering
\caption{Comprehensive Performance Summary. G22 data are taken at $T=1000$, with 10 independent repeated experiments performed. The G1–G81 data represent aggregated statistics from 71 graphs. Bold indicates the optimal value in each column. *Contains multiple startup strategies.}
\label{tab:summary} 
\scriptsize
\begin{adjustbox}{max width=\textwidth}
\begin{tabular}{lcccccc}
\toprule
Algorithm & G22 Gap$_{\text{mean}}$ & G1-G81 Gap median & AR Mean & Gap-best graphs & AR-best graphs & TTS$_{0.99}$ median (s) \\
\midrule
Standard BSB & 0.33\% & 1.23\% & 0.915 & 0 & 0 & 151.3 \\
Standard DSB & 0.68\% & 0.78\% & 0.920 & 7 & 2 & \textbf{2.3} \\
\textbf{ME-BSB} & 0.26\% & \textbf{0.66\%} & \textbf{0.991} & 17 & \textbf{28} & \textbf{2.1} \\
SE-DSB & \textbf{0.04\%}$^*$ & 1.01\% & 0.943 & \textbf{25}$^*$ & \textbf{28}$^*$ & 11.4$^*$ \\
SG-DSB & \textbf{0.04\%}$^*$ & 1.67\% & 0.931 & 11$^*$ & 4$^*$ & 13.0$^*$ \\
GSB-BSB & 0.40\% & 1.26\% & 0.977 & 2 & 2 & 43.1 \\
GSB-DSB & 0.34\% & 1.41\% & 0.978 & 2 & 0 & 53.5 \\
Tabu-BSB & 0.48\% & 1.65\% & 0.959 & 2 & 2 & 5877.5 \\
Tabu-DSB & 0.62\% & 1.51\% & 0.960 & 5 & 5 & 100.1 \\
\bottomrule
\end{tabular}
\end{adjustbox}
\end{table}

Table~\ref{tab:summary} gives the cross-level comparison. On G22, SE-DSB and SG-DSB both achieve gaps below 0.05\%, tied for the best. Across the 71 graphs of G1--G81, the picture shifts: ME-BSB delivers the lowest median gap (0.66\%) and the highest mean AR (0.991), and ties with SE-DSB for the most AR-best graphs (28 each). SE-DSB leads in Gap-best graphs with 25, but its AR mean is dragged down considerably by the frustration-free graphs---once G1--G5 are set aside, its median gap sits around 1.0\%, comparable to ME-BSB. SG-DSB's cross-graph median gap is notably higher than SE-DSB's; the side effects of density-aware scheduling and velocity momentum on a few difficult graphs likely contribute to this. The GSB variants post competitive AR means, but their TTS and winner counts are far behind AE-QSB. Tabu-BSB's median TTS of 5877.5~s has been discussed earlier and will not be revisited here.

\subsection{Ablation findings}
\label{sec:ablation-findings}

Ablation experiments were run at two tiers---single-factor and multi-factor combinations (design in \secref{sec:setup}, full data in \ref{app:ablation})---and they give a clear ranking of the components by importance.

The strongest finding concerns the exploration subpopulation. In both ME-BSB and SG-DSB, removing it alone sends Gap up by roughly 7.5 percentage points, a degradation far worse than that of any other component. The rescue mechanism comes second: removing it causes sustained deterioration of tail quality, with Gap rising by about 0.14 percentage points. More telling is the multi-factor result: the $(-\text{exploration}, -\text{rescue})$ combination produces super-additive degradation---the combined loss significantly exceeds the sum of the two individual losses. The exploration subpopulation keeps diversity high enough that rescue operates under light load; once exploration is turned off and the population collapses, rescue gets triggered repeatedly and becomes the only repair channel, and this imbalance shows up as super-additive loss.

Density-aware adaptive scheduling is what sets SG-DSB apart from SE-DSB. On G22 alone its independent contribution is modest (roughly 0.02 percentage points), but on the extremely dense segments G31--G40, SG-DSB with density awareness nearly covers SE-DSB without it---its value is in cross-graph generalization, not single-graph refinement. Velocity momentum contributes almost nothing on its own under the G22 setup ($\Delta$Gap $\approx$ 0). Several ablation controls agree that the exploration subpopulation and density-aware scheduling already handle the search stability that momentum is meant to provide. It may prove more useful at larger scales or under higher noise.

ME-BSB's ablation runs in the opposite direction and makes the case for simplicity. Removing F-switch, Tabu repulsion, and diversity restart each yields a slight improvement of 0.05--0.10 percentage points. Under the single-elite plus 15\% exploration subpopulation configuration, these auxiliary mechanisms are already covered by the base dynamics; keeping them adds complexity without benefit. Full data and the component importance ranking appear in \ref{app:ablation} (Figs.~A1, A2).

\subsection{Complementarity and population sensing}
\label{sec:complementarity}

Taken together, the G22 and G1--G81 results map out fairly clear boundaries and complementarity among the three algorithms. ME-BSB, with single-elite guidance and the exploration subpopulation, delivers strong extreme-value capability and the shortest runtime, making it a natural fit for latency-sensitive settings. SE-DSB replaces hard switching with linear mixed coupling and drives G22's gap down to 0.04\%, confirming that smooth transitions bring real precision gains. SG-DSB layers density-aware scheduling on top and nearly covers SE-DSB on the dense segments.

The segmentation analysis in G1--G81 makes this complementarity more systematic. On the easy frustration-free graphs (G1--G5), GSB and Tabu variants still perform best---constant strong guidance converges fast on simple landscapes. Once the frustrated spin-glass segment begins (G6--G10), AE-QSB wins every graph. The pattern continues: sparse frustration-free (G14--G17) goes entirely to ME-BSB; mixed-sign segments (G11--G13, G18--G20) go to SE-DSB/SG-DSB; near-optimal (G21--G26) and truly hard dense segments (G27--G30) are likewise covered by the two. On the extremely dense segments (G31--G34, G39--G40), SG-DSB's density awareness kicks in; on the mid-dense segment (G41--G50), ME-BSB dominates overwhelmingly. At the very large scale (G71--G81), the winner shifts back to Tabu-DSB. A clean pattern emerges: as graph difficulty rises, the best algorithm migrates from Tabu/GSB through ME-BSB and SE-DSB/SG-DSB, and back to Tabu-DSB---each covers the others' weak spots.

This brings us back to the paper's basic claim: does closed-loop control driven by $(D, F, Q, R)$ actually work? Three lines of evidence say yes. Conceptually, Section~3 argued that the four indicators are complementary without being redundant. Experimentally, G22 and G1--G81 consistently show that adding adaptive control lifts performance well above the fixed-schedule baselines. In terms of generalization, using only coarse-grained segmentation---no per-graph tuning---already wins lowest gap on 74.6\% of the graphs, which means the four indicators do capture the information that matters across different graph classes and evolution stages.

\subsection{Efficiency, limitations, and outlook}
\label{sec:efficiency}

On the efficiency side, SG-DSB's per-run cost is slightly higher than SE-DSB's, with the increment coming mainly from rescue condition checks and momentum EMA updates; multi-launch total time scales roughly linearly with the number of launches. ME-BSB runs noticeably faster in single-run mode and suits latency-sensitive applications. The sensing layer fires once every 50 steps, so its overhead amortizes to about $O(NB/50)$ per step---not a bottleneck.

As for parameters, each of the three algorithms carries a dozen to over twenty tunable values. Ablation has already shown that quite a few of them are not independently effective: Tabu repulsion, diversity restart, and the multi-elite pool under ME-BSB yield slight improvements when removed; velocity momentum under the G22 setup produces almost no measurable effect. Pruning these redundant parameters is the next step. But to fundamentally reduce the reliance on manual tuning, automatic parameter selection based on graph structural features is needed---and that remains an open direction.

Adaptive bitflip achieves a workable balance between efficiency and terminal quality, though the per-column recomputation of $J \cdot \mathbf{s}$ ($O(\text{nnz})$) means refinement cost grows linearly with edge count. Once $N$ exceeds $10^4$, Hamming-distance computation and per-column rescue also begin to take meaningful time. On the theoretical side, our analysis of the four indicators' independence is currently constructive; a tighter characterization would strengthen the framework's foundation.

In short, the core idea behind AE-QSB---using lightweight runtime state sensing to turn fixed scheduling into closed-loop adaptive control---provides a technical route that has been verified to work for SB-type algorithms. Whether this sensing-driven control approach generalizes to other population-based parallel evolution frameworks is a question for follow-up work.

\section{Conclusion}
\label{sec:conclusion}

This paper started from three characteristic problems of existing SB methods under fixed scheduling---population collapse from uniform guidance, scheduling mismatch from rigid phase division, and manual tuning of noise mechanisms---and proposed the AE-QSB framework. The framework's core is to use four population statistics $(D, F, Q, R)$ for real-time state sensing and, based on the sensed state, to adjust step sizes, coupling modes, guidance strength, and restart strategies online, forming a complete sense-decide-execute loop. Three algorithm instances were built under this framework, each with a different emphasis: ME-BSB uses $F$-driven hard switching with a weak exploration subpopulation to push extreme values (Gap 0.26\% on G22, shortest runtime); SE-DSB uses linear mixed coupling to achieve population-wide uniform refinement (Gap 0.04\%); SG-DSB further adds density-aware scheduling, which strengthens cross-graph generalization. Across 71 heterogeneous G-set instances, the three AE-QSB variants together achieved the lowest mean gap on 74.6\% of graphs and the highest AR on 84.5\%. The three algorithms complement each other along the difficulty spectrum---from Tabu/GSB on simple graphs, through AE-QSB variants on frustrated and dense graphs, to Tabu-DSB at very large scale---with clean, coherent boundaries.

Ablation experiments (30 variants) established a clear ordering. The exploration subpopulation ranks first: its removal sends Gap up by roughly 7.5 percentage points and it is irreplaceable under any configuration. The rescue mechanism ranks second: its individual removal causes far less damage than losing exploration, but the two together exhibit super-additive degradation, confirming functional coupling between them. Density-aware scheduling is the key gain that lets SG-DSB generalize to heterogeneous benchmarks, as borne out by its performance on the extremely dense segments. Velocity momentum contributes little under G22; its function is covered by the exploration subpopulation and density-aware scheduling. ME-BSB's ablation confirms the value of simplicity more directly: removing F-switch, Tabu repulsion, and diversity restart each brought a slight performance gain, indicating that under the single-elite plus exploration configuration, these components are redundant.

Several limitations and directions are worth noting. The number of parameters remains high; ablation has helped trim some redundancies, but building graph-feature-based automatic parameter selection is the more fundamental solution. Bitflip refinement cost grows linearly with edge count, and beyond $N > 10^4$, Hamming-distance computation and rescue operations also become bottlenecks. The independence argument for the four indicators is currently constructive and would benefit from a more rigorous theoretical treatment. Finally, whether AE-QSB's sensing-driven closed-loop approach can be adapted to other population-based parallel evolution frameworks is a question that needs further experiments to answer.

\Acknowledgements{This work was supported by the National Natural Science Foundation of China (Grant Nos. 62271070 and 62176273).}

\bibliographystyle{scis}
\bibliography{ref}

\appendix
\section{G22 Ablation Experiments}
\label{app:ablation}

This appendix gives the full ablation results for ME-BSB and SG-DSB on G22 ($N=2000$, frustrated graph, $T=1000$, $B=256$). The experimental setup follows \secref{sec:setup}: two-tier ablation design, each variant only changes the target component's enable flag, all other parameters unchanged, 5 independent repeats.

\subsection{Ablation design}

\textbf{Single-factor ablation (tier 1).} Starting from the complete algorithm, a single mechanism is removed or enhanced at a time. ME-BSB family: 7 variants---remove F-switch, remove exploration subpopulation, remove Tabu repulsion, remove diversity emergency restart, add multi-elite pool ($K=6$), add postprocessing. SG-DSB family: 9 variants---remove velocity momentum, remove density-aware adaptive scheduling, remove exploration subpopulation, remove Tabu repulsion, remove diversity emergency restart, remove rescue mechanism, add multi-elite pool ($K=5$), add postprocessing. SE-DSB (static-enhanced baseline without momentum) and SG-DSB (legacy) (the older multi-elite $+$ postprocessing version) serve as architectural baselines.

\textbf{Multi-factor combination ablation (tier 2).} Two or three mechanisms are removed together to check for cooperation or cancellation effects. ME-BSB family: 4 combinations: $(-F_{\text{switch}}, -\text{exploration})$, $(-\text{tabu}, -\text{restart})$, $(-F_{\text{switch}}, -\text{tabu})$, $(-F, -\text{explore}, -\text{tabu})$. SG-DSB family: 5 combinations: $(-\text{momentum}, -\text{density})$, $(-\text{momentum}, -\text{rescue})$, $(-\text{exploration}, -\text{rescue})$, $(-\text{tabu}, -\text{restart})$, $(-\text{mom}, -\text{density}, -\text{rescue})$.

\subsection{Interaction effect analysis}

We start with the two most notable combinations (core data in Figs.~A1 and A2; exact values fluctuate slightly across runs, but relative component ordering and interaction patterns are stable).

ME-BSB's $(-F_{\text{switch}}, -\text{exploration})$ combination shows sub-additivity: the improvement from removing F-switch partially offsets the degradation from losing the exploration subpopulation. These two mechanisms operate at different stages of search, so removing both does not simply sum their individual losses. SG-DSB's $(-\text{exploration}, -\text{rescue})$ combination goes the other way---super-additive degradation, where the combined loss clearly exceeds the sum of the two individual losses. The reason is straightforward: with exploration active, population diversity stays high and rescue operates under light load. Once exploration is turned off, the population collapses quickly, rescue gets triggered repeatedly, and becomes the only repair channel. The individual loss from removing rescue looks small because exploration is still there to back it up; remove both and that backup is gone, which amplifies the loss.

Next, functional redundancy. Velocity momentum smooths the guidance direction with EMA, which is meant to improve search stability. But the exploration subpopulation and density-aware adaptive scheduling already cover the same need from different angles. Several ablation controls agree on one point: under the G22 setup, removing velocity momentum leaves Gap essentially unchanged. This does not rule out independent value at larger scales or under higher noise, but under the current experimental conditions its contribution is indeed covered by the other components. The triple-stripping results agree: after SG-DSB removes momentum, density-aware scheduling, and rescue together, the degradation mainly comes from the rescue loss; ME-BSB's triple-stripping degradation is small because its performance core is the early exploration efficiency of BSB continuous coupling and the self-stabilizing property of diversity-gated elite guidance. The peripheral components have limited impact.

\subsection{Component importance ranking}

Figures~A1 and A2 summarize all 30 ablation variants from two complementary angles. Fig.~A1 is a four-panel composite: (a)--(c) use horizontal bars to show $\Delta$Gap$_{\text{mean}}$ relative to the complete algorithm (blue = improvement, red = degradation), and (d) scatters Gap$_{\text{mean}}$ against runtime. Fig.~A2 is a four-dimensional bar chart with all variants sorted by ascending Gap$_{\text{mean}}$, comparing Gap$_{\text{mean}}$, Gap$_{\text{max}}$, AR$_{\text{mean}}$, and runtime side by side; the Standard BSB baseline is marked with a dashed line as a lower-bound reference.

\begin{figure}[!t]
\centering
\aeqsbincludegraphics[width=0.95\textwidth]{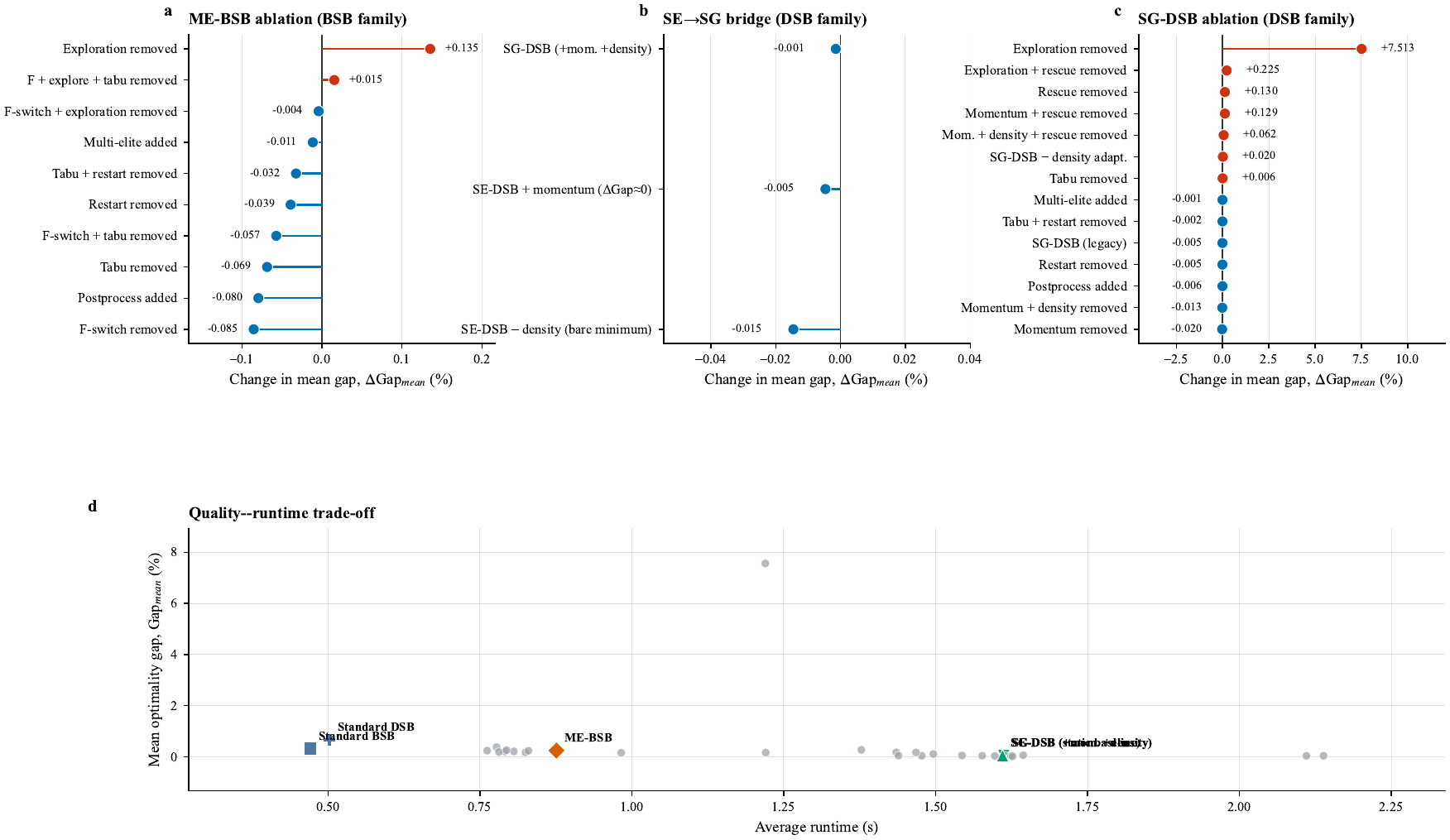}
\caption{Ablation composite figure. (a) ME-BSB single-factor ablation. (b) SE-DSB $\to$ SG-DSB bridging. (c) SG-DSB single-factor ablation, multi-factor variants in darker tones. (d) Gap$_{\text{mean}}$ vs.\ runtime scatter; complete algorithms shown with distinct markers, ablation variants as circles.}
\label{fig:ablation-composite}
\end{figure}

\begin{figure}[!t]
\centering
\aeqsbincludegraphics[width=0.95\textwidth]{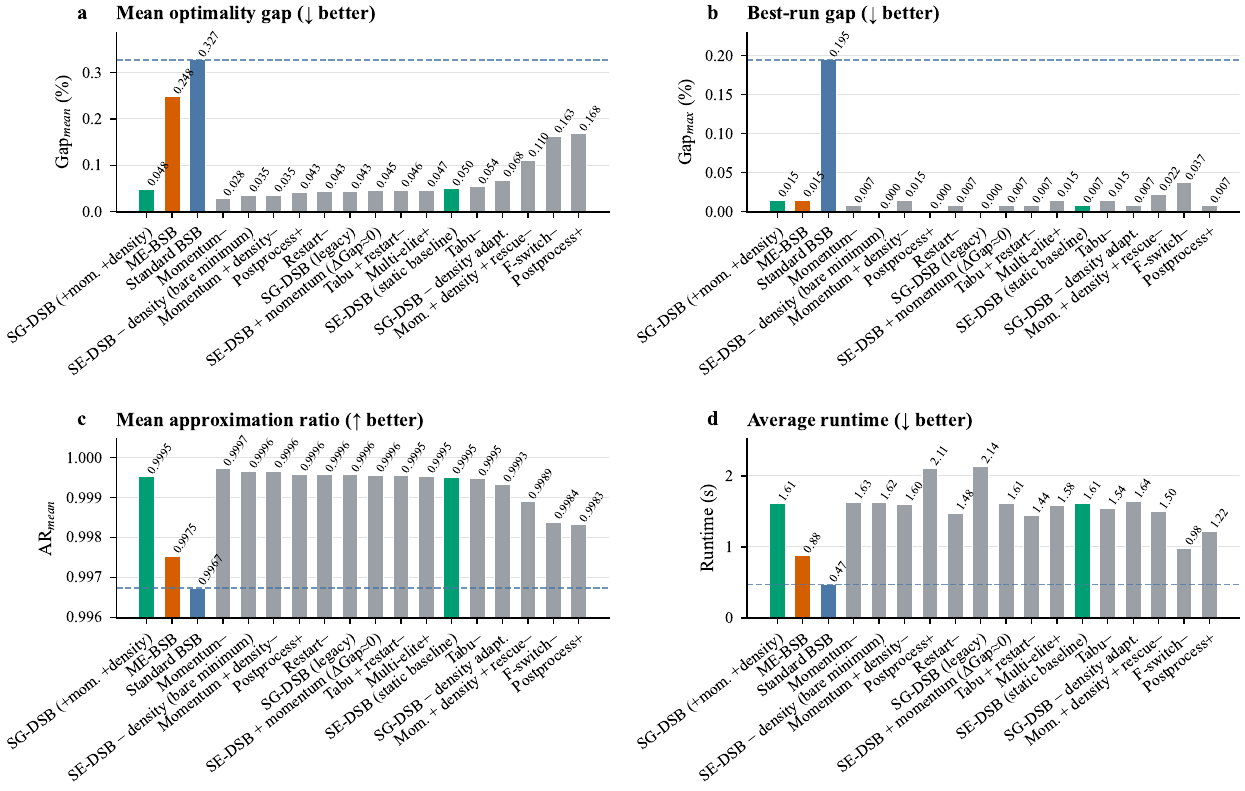}
\caption{Ablation multi-metric comparison. All variants compared on four metrics (sorted by ascending Gap$_{\text{mean}}$). Standard BSB baseline as dashed line. Lower Gap and runtime are better; higher AR is better.}
\label{fig:ablation-metrics}
\end{figure}

Taken together, the two figures give the following component ranking.

\textbf{The exploration subpopulation ranks first.} Its removal sends Gap up by roughly 7.5 percentage points, far exceeding the effect of removing any other single component, and this degradation appears consistently across all four quality dimensions in Fig.~A2. The result holds for both ME-BSB and SG-DSB---regardless of architecture, keeping an unguided independent search channel is the most basic requirement.

\textbf{The rescue mechanism ranks second.} Its individual removal raises Gap by about 0.14 percentage points, far below the exploration subpopulation. But as noted in Section~A.2, rescue and exploration are functionally coupled: the combined removal amplifies the loss to roughly 7.7\%, well beyond the sum of the two individual losses.

\textbf{Density-aware adaptive scheduling} is what distinguishes SG-DSB from SE-DSB. On G22 its gain is modest (roughly 0.02 percentage points), but it is fully unleashed on the dense segments G31--G40 (see \secref{sec:g1g81}), indicating that its value lies in cross-graph generalization rather than single-graph refinement.

\textbf{ME-BSB's three auxiliary mechanisms are redundant.} Removing F-switch, Tabu repulsion, and diversity restart each \emph{lowers} Gap by 0.05--0.10 percentage points. Under the single-elite plus 15\% exploration configuration, the base dynamics already cover what these mechanisms provide; keeping them only adds complexity.

\textbf{Velocity momentum contributes nothing on its own under G22} ($\Delta$Gap $\approx$ 0). Several ablation controls consistently show that its function is fully compensated by the exploration subpopulation and density-aware adaptive scheduling.

\end{document}